\documentclass[lettersize,journal]{IEEEtran}
\usepackage{amsmath,amsfonts}
\usepackage{algorithmic}
\usepackage{algorithm}
\usepackage{array}
\usepackage[caption=false,font=normalsize,labelfont=sf,textfont=sf]{subfig}
\usepackage{textcomp}
\usepackage{stfloats}
\usepackage{url}
\usepackage{verbatim}
\usepackage{graphicx}
\usepackage{cite}
\begin{document}

\title{Learning Variable Impedance Skills from Demonstrations with Passivity Guarantee}

\author{Yu Zhang, Long Cheng, Haoyu Zhang, Xiuze Xia, Rongtao Xu, Yongxiang Zou, and Houcheng Li,
\thanks{*This work was supported in part by the National Natural Science Foundation of China (Grants 62025307, U1913209) and was also supported by the CAS Project for Young Scientists in Basic Research (Grant No. YSBR-034)}
\thanks{The authors are all with the School of Artificial Intelligence, University of Chinese Academy of Sciences, Beijing 100049, China. They are also with the State Key laboratory of Multimodal Artificial Intelligence Systems, Institute of Automation, Chinese Academy of Sciences, Beijing 100190, China. All correspondences should be addressed to the corresponding author Dr. Long Cheng (long.cheng@ia.ac.cn).}}

\markboth{Journal of \LaTeX\ Class Files,~Vol.~14, No.~8, August~2021}%
{Shell \MakeLowercase{\textit{et al.}}: A Sample Article Using IEEEtran.cls for IEEE Journals}


\maketitle

\begin{abstract}

Robots are increasingly being deployed not only in workplaces but also in households. The effective execution of manipulation tasks by robots relies on variable impedance control, which is essential for managing contact forces. Moreover, robots should possess adaptive capabilities to handle the significant variations inherent in different tasks within dynamic environments, capabilities that can be acquired through human demonstrations.
This paper presents a learning-from-demonstration framework that integrates force sensing and motion information to enhance variable impedance control. The proposed algorithm involves estimating full stiffness matrices from human demonstrations, which are then combined with motion data to create a model using a non-parametric method. This model enables the robot to replicate demonstrated tasks and to adapt to new conditions through a state-dependent stiffness profile. Additionally,  A novel Lyapunov function is designed to ensure system passivity by leveraging the learned stiffness parameters to achieve variable impedance control. The proposed approach was evaluated using both virtual simulations of variable stiffness systems and practical experiments with a real robot. The first evaluation demonstrates that the stiffness estimation method exhibits superior robustness compared to traditional techniques. The second evaluation shows that the proposed framework is straightforward to implement, effectively enabling variable impedance control.
\end{abstract}


\section{Introduction}
\IEEEPARstart{T}{he} use of robots has become increasingly widespread in various fields, including manufacturing and rehabilitation \cite{b1,b2}. This is largely due to their ability to perform complex manipulation tasks in unknown and unstructured environments where fixed coding is impractical.  Learning from demonstration (LfD) is a suitable approach for these scenarios, as it is an intuitive and user-friendly method that enables robots to implicitly learn task constraints and acquire manipulation skills through demonstrations \cite{b3}. This approach eliminates the need for hard coding different tasks, making robots more adaptable and efficient.

Most LfD methods primarily focus on learning motion trajectories \cite{b4,ralzy}. However, for robots operating in unknown and unstructured environments, learning motion trajectories alone is not sufficient. It is crucial to extend robot learning capabilities to include force and impedance domains \cite{b6,b7}, which are essential for robots working in dynamic environments.

\begin{figure}[htbp]
	\centering
	\includegraphics[width=0.5\textwidth]{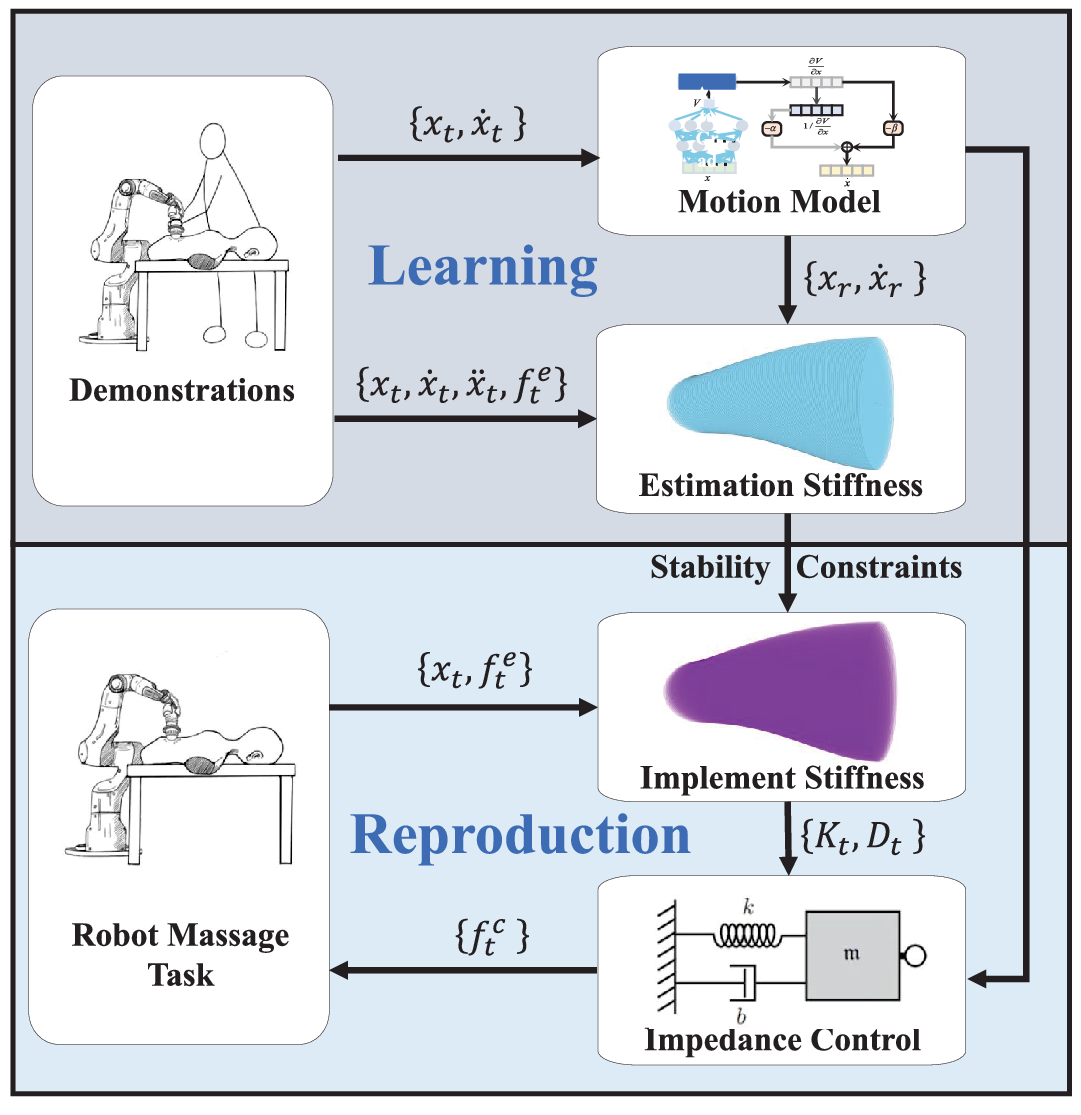}		
	\caption{The framework involves learning variable stiffness from demonstrations and implementing variable impedance control on the Franka robot to perform a massage task.}		
	\label{fig1}
\end{figure}
This paper proposes a framework (shown in Fig. \ref{fig1}) that learns variable impedance skills from demonstrations while ensuring passivity. The proposed learning framework estimates stiffness from demonstrations and implements variable impedance control using a stability condition derived from a novel Lyapunov approach that guarantees passivity with the learned stiffness. The main contributions of this paper are listed as follows:
\begin{itemize}
	\item A effectiveness stiffness estimating method is proposed that can estimate the demonstrate stiffness from the demonstrations, regardless of  whether the damping information is known in advance.
	
	\item A novel Lyapunov function is proposed to derive an easily implementable stability condition for variable impedance control. The learned stiffness can then be used to achieve the task as demonstrated by the human.
	
	\item The experimental results provide conclusive evidence of the robustness and practicality of the proposed framework for learning stiffness from demonstrations and utilizing the learned stiffness for variable impedance control in real-world experiments.
\end{itemize}
The paper is structured as follows: Section \ref{sec2} provides an overview of the current state of research in the field. Section \ref{sec3} introduces the proposed approach for effective stiffness estimation, detailing how to integrate this estimated stiffness with motion information to develop a comprehensive model. This section also covers the theoretical foundations of the proposed stability conditions for variable impedance control. Section \ref{sec4} presents the results of both simulations and real robot experiments, which validate the performance of the proposed methods.  Finally, Section \ref{sec5} concludes with a summary of the key findings and contributions of the proposed framework.
\section{RELATED  WORK}\label{sec2}
Humans possess an exceptional ability to adapt their stiffness when interacting with complex environments during manipulation tasks. This adaptability allows them to effectively control their level of stiffness or resistance based on the specific requirements of the task and the characteristics of the environment. Previous research has focused on understanding how humans modulate impedance while interacting with their environment and transferring these skills to robots \cite{b8,b9}. However, previous robot learning capabilities did not include automatic modification of impedance control parameters to respond to unforeseen situations. Recently, there has been a surge of interest in robot learning approaches for modeling variable impedance skills, in tasks where a robot needs to physically interact with the environment, including humans.

Researchers have developed various methods for generating variable stiffness parameters for the controller based on either task rules, movement models derived from human demonstration, or reinforcement learning \cite{b10,b11,b12,b13}.  Based on the rule of assist-as-need, \cite{b14}  utilizes the electromyography (sEMG) signals from muscles related to the joint to estimate both joint torque and quasi-stiffness of human in order to map to the voluntary efforts of the human subject. Then the manually designed control strategy adjusts the degree of assistance by varying the stiffness for variable impedance control. Specifically, When the robot detects a need to assist the human user, the stiffness parameter of the controller is set to a higher value, on the other hand, when the system detects that the human user is attempting to move actively,
the stiffness parameter is set to be a lower value. In \cite{b15}, a controller for coupled human-robot systems is proposed to improve stability and agility while reducing effort required by the human user. The controller achieves this goal by dynamically adjusting both damping and stiffness. Specifically, the controller applies robot damping within a range of negative to positive values to add or dissipate energy depending on the intended motion of the user.  The authors in \cite{stifftro,ras,emg2,emgtm,myras}, developed a variable impedance controller that uses the time-varying stiffness parameters estimated from human demonstrations.   By utilizing the estimated stiffness parameters, the controller allows the robot to adapt its impedance or stiffness during interactions, mimicking the natural and intuitive behavior observed in humans. 
In \cite{rl}, the authors propose a method for learning a policy that enables robots to generate behaviors in the presence of contact uncertainties. The approach involves learning two key components: the output impedance and the desired position in joint space. Additionally, an additional reward term regularizer is introduced to enhance the learning process. The output impedance refers to the robot's ability to modulate its interaction forces and stiffness during contact with the environment. By learning the appropriate output impedance, the robot can effectively adapt to different contact uncertainties, ensuring stable and controlled interactions.

The preceding studies demonstrated various methods for designing variable stiffness profiles that enable robots to perform different interaction tasks. The variable impedance controller enhances the robot's ability to interact with its environment by dynamically adjusting its stiffness based on the task requirements and environmental conditions. However, it has been observed that the use of variable stiffness parameters in impedance controllers can potentially compromise the stability of physical human-robot interactions \cite{stabilitytro, tank}. 
The passivity theory, which is widely used in the field of robotics, provides an energy-based perspective for evaluating the coupled stability of physical human-robot interaction. This theory allows for an analysis of the energy flow and exchange between the robot and the human, offering a valuable framework for designing robot behaviors that are both safe and effective \cite{b24}. In the context of physical human-robot interaction, the passivity of a robot-human system is determined by comparing the energy stored in the robot with the external energy injected by the environment, which is typically attributed to the human. A passive system is characterized by the condition that the energy stored in the robot remains consistently lower than the external energy injected by the environment. This definition of passivity ensures that the robot does not generate additional energy and remains stable even in the absence of any external forces. By maintaining a lower energy level than that injected by the human, the robot avoids becoming energetically dominant and guarantees that the interaction remains safe and controlled.
In \cite{stabilitytro}, the authors introduce a novel Lyapunov candidate function that offers more relaxed stability conditions for a system, which potentially expands the range of system behaviors that can be considered stable. However, the proposed conditions may not be sufficient for robots operating in unstructured environments. To ensure the stability and robustness of robots operating in unstructured environments, additional considerations and techniques may be necessary. The authors in \cite{tank} introduce a virtual energy-storing tank that can absorb and store the dissipation energy from the robot. When the virtual tank is available, it compensates for any additional energy generated by the robot and provides a more straightforward and practical  way for preserving the passivity of the robot-human system. By effectively compensating for the excess energy generated by the robot, the virtual tank ensures that the energy balance within the system is maintained. This compensation mechanism simplifies the design and control of the robot, as it provides a approach to regulate and manage the energy flow without compromising the passivity requirements.

\section{Proposed Approach}\label{sec3}

\begin{figure}[htbp]
	\centering
	\includegraphics[width=0.5\textwidth]{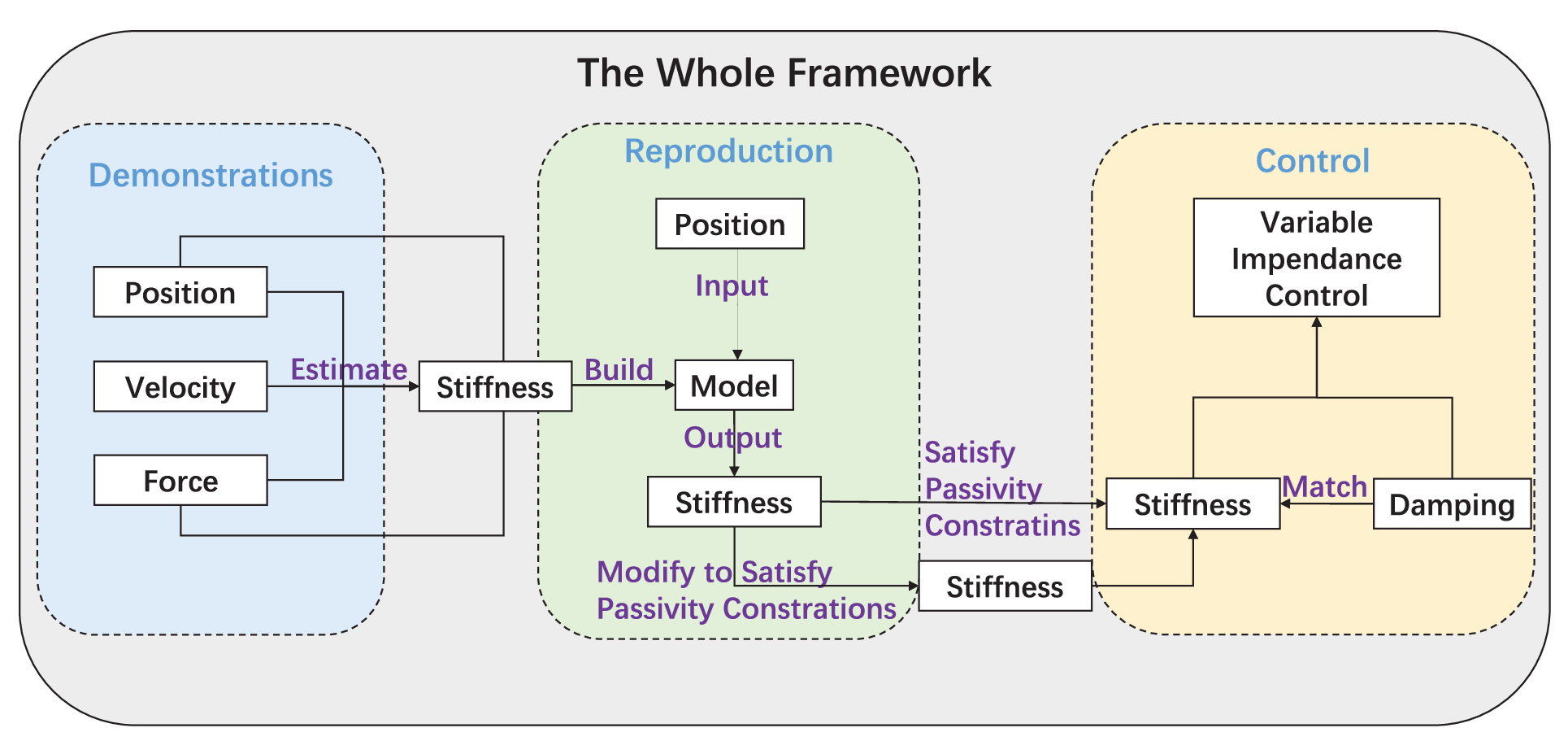}		
	\caption{The process for extracting variable stiffness parameters from demonstration data, followed by the application of these parameters in the implementation of variable impedance control, ensuring the system maintains passivity.}		
	\label{fig2}
\end{figure}

This section introduces a novel architecture for LfD that not only facilitates the learning and replication of variable impedance skills in cooperative tasks but also ensures passivity. Incorporating passivity into the architecture is critical because when transferring human's variable stiffness skills, it ensures that the robot can effectively learn and replicate these skills while maintaining stability and safety. 

\subsection{Interaction Model}\label{sec3_1}
Human beings demonstrate a remarkable ability to apply appropriate force to achieve desired dynamic relationships with their environment. This capacity finds a counterpart in impedance control, a highly effective method for regulating interaction behaviors between robots and their working environment. By learning from human demonstrations,  robots can acquire the ability to perform tasks that require adaptive impedance control.

Given that the majority of interactions take place within the task space, they predominantly involve the utilization of the robots' end-effector. Specifically, the movement of a robot's end-effector within the task space can be effectively represented as a mass unit influenced by two primary factors: interaction forces denoted as $\boldsymbol{f}^{e}$, and the control input denoted as $\boldsymbol{f}^{c}$. This modeling approach is supported by \cite{ok87}:
\begin{equation}\label{equ:1}
	\boldsymbol{H}\ddot{\boldsymbol{x}} = \boldsymbol{f}^{e} + \boldsymbol{f}^{c},
\end{equation}
where $\ddot{\boldsymbol{x}}$ denotes the acceleration of the motion and $\boldsymbol{H}$ denotes a positive definite matrix that remains unchanged for the task. By configuring the motion control forces as $\boldsymbol{f}^{c}$ in the presence of an interaction force $\boldsymbol{f}^{e}$,  the desired dynamics can be achieved. 

Impedance control allows the modeling of the robot's motion during interaction as a virtual mass-spring-damper (MSD) system at each time step $t$. Specifically, assuming that the reference motions are available, which can be designed manually or learned from demonstrations, the equation of 
impedance motion can be expressed as follows:
\begin{equation}\label{equ:2}
	\boldsymbol{K}_{t}(\boldsymbol{x}_{t}-\boldsymbol{x}^{r}_{t})+\boldsymbol{D}_{t}(\dot{\boldsymbol{x}}_{t} -\dot{\boldsymbol{x}}^{r}_{t})+\boldsymbol{H}(\ddot{\boldsymbol{x}}_{t}-\ddot{\boldsymbol{x}}^{r}_{t}) =\boldsymbol{f}_{t}^{e},
\end{equation}
where  $\boldsymbol{x}_{t}^{r}$, $\boldsymbol{\dot{x}}_{t}^{r}$ $\boldsymbol{\ddot{x}}_{t}^{r}$, $\boldsymbol{K}_{t}$, and $\boldsymbol{D}_{t}$,  represent position, velocity, acceleration of the reference trajectory, as well as the full stiffness matrix, damping matrix for the virtual MSD system, respectively. Conversely, $\boldsymbol{x}_{t}$, $\dot{\boldsymbol{x}}_{t}$, $\ddot{\boldsymbol{x}}_{t}$, and $\boldsymbol{f}_{t}^{e}$ denote the position, velocity, acceleration, and interaction forces of the real motion, respectively.

$\dot{\boldsymbol{x}}_{t}$ and $\ddot{\boldsymbol{x}}_{t}$ are the first and second time derivatives of ${\boldsymbol{x}}_{t}$ that can be easily computed once the position and time information are acquired, and the interaction forces $\boldsymbol{f}_{t}^{e}$ can be directly measured using a force sensor mounted on the robot's end-effector.
By using $\boldsymbol{e}, \dot{\boldsymbol{e}}, \ddot{\boldsymbol{e}}$ to represent the position, velocity, and acceleration errors between real motions and reference motions, equation (\ref{equ:2}) can be reformulated as:
\begin{equation}\label{equ:3}
	\boldsymbol{K}_{t}\boldsymbol{e}_{t}+\boldsymbol{D}_{t}\dot{\boldsymbol{e}}_{t} +	\boldsymbol{H}\ddot{\boldsymbol{e}}_{t} =\boldsymbol{f}_{t}^{e}.
\end{equation}
Subsequently, the robot's behavior can be customized by modifying the stiffness within the refer trajectories in the interaction model (\ref{equ:2}). To attain an ideal human-like behavior, the variables  $\boldsymbol{K}_{t}$ can be acquired through demonstrations conducted by human teachers. And In the context of human-robot interaction, the robot's slow movement allows us to simplify the dynamics of the system. Specifically, equation (\ref{equ:3}) can be approximated as a second-order damping-spring system, described as follows:
\begin{equation}\label{equ:sm1}
	\boldsymbol{K}_{t}\boldsymbol{e}_{t}+\boldsymbol{D}_{t}\dot{\boldsymbol{e}}_{t} + 	\boldsymbol{H}\ddot{\boldsymbol{x}}_{t} =\boldsymbol{f}_{t}^{e}.
\end{equation}

By observing and recording the actions and behaviors of skilled human teachers, the robot can learn and extract the desired variable parameters $\boldsymbol{K}_{t}$ that represent the stiffness characteristics of the human.

\subsection{Stiffness Estimation from Demonstrations}\label{sec3_2}
To obtain the full stiffness matrix $\boldsymbol{K}_{t}$, experts have explored various methods. One approach utilized by \cite{stifftro} involves the use of the Gaussian mixture model (GMM) to cluster the demonstrated data and estimated $\boldsymbol{K}_{t}$ by obtaining the inverse of the position covariance. However, this approach requires multiply demonstrations and cannot be applied to tasks where the robot's end-effector is constrained to follow a single path.
Another approach, as described by \cite{emg2} utilized the electromyography (sEMG) signals to estimate stiffness.
This approach calculates the activation level of human muscles and offers the advantage of being cost-effective while providing real-time estimation. In a similar vein,
\cite{emgtm}  specifically emulates imitates the underlying the inherent  transformation within
the muscular and skeletal systems. They employs the model-based approach to estimate the joint torque and joint stiffness simultaneously without additional calibration procedures. However, the sEMG based approaches are highly sensitive to noise and individual differences.  

A different approach proposed by \cite{myras} uses the neural network to estimate the stiffness and damping from the demonstrations, based on the states of end-effector and the interaction force. Although this method can easily generalize to different states,  training the neural network is time consuming and the accuracy is not always satisfactory. In contrast,
\cite{ras} proposes an algorithm that uses the least squares method with the nearest positive definite matrix (SPD) method to estimate the stiffness with 
a sliding window at each time step quickly. However,  a predetermined damping value is acquired by this algorithm, which is not known in the real demonstrations. Therefore, the accuracy of the estimated stiffness relies on the empirically designed value of the damping. Inspired by \cite{ras}, a novel stiffness estimation algorithm that inherit the highly effectiveness of \cite{ras} while do not need to know the damping is proposed in this paper. 

To introduce the proposed stiffness estimation algorithm, the algorithm proposed in \cite{ras} is introduced firstly. In human demonstration, with the reference trajectories, the presented data in the demonstrations include position, velocity, acceleration, and external force vectors, respectively.
Then to estimate the full stiffness matrix at each time step $t$, \cite{ras} employs a sliding window $L$ that moves along the demonstration data.  The small sliding window is employed to treat stiffness as an invariant over a short period of time.

Equation
(\ref{equ:sm1}) can be also expressed as:
\begin{equation}\label{equ:4}
	\boldsymbol{K}_{t}\boldsymbol{e}_{t}=  \boldsymbol{f}_{t}^{e}-\boldsymbol{H}_{t}\boldsymbol{\ddot{x}}_{t}-\boldsymbol{D}_{t}\boldsymbol{\dot{e}}_{t}.
\end{equation}
By defining $\boldsymbol{f}_{t}^{e}-\boldsymbol{H}_{t}\boldsymbol{\ddot{x}}_{t}-\boldsymbol{D}_{t}\boldsymbol{\dot{e}}_{t}$ as $\tilde{\boldsymbol{y}}_{t}$ and $\boldsymbol{e}_{t}$ as $\tilde{\boldsymbol{x}}_{t}$, equation (\ref{equ:4}) becomes:
\begin{equation}\label{equ:5}
	\boldsymbol{K}_{t}\tilde{\boldsymbol{x}}_{t}=\tilde{\boldsymbol{y}}_{t}.
\end{equation}
Hence, a rough approximation stiffness $\boldsymbol{\hat{B}}_{t}$ at each step can be calculated by using the least squares method. Subsequently, a nearest-SPD matrix is calculated to represent the estimated stiffness.
To calculate the nearest-SPD matrix,  $\boldsymbol{B}_{t}=\frac{\boldsymbol{\hat{B}}_{t}+\boldsymbol{\hat{B}}^{\mathrm{T}}_{t}}{2}$ is first computed. Next, $\boldsymbol{B}_{t}$ is decomposed using a polar decomposition as
$\boldsymbol{B}_{t}=\boldsymbol{U}_{t}\boldsymbol{L}_{t}$. Finally, the estimated stiffness is represented as $\boldsymbol{\hat{K}}_{t}=\frac{\boldsymbol{B}_{t}+\boldsymbol{L}_{t}}{2}$,  then the estimated stiffness is a $N$ dimensional symmetric and positive definite matrix \cite{spd}.

It can be seen that to estimate the full stiffness matrix from the demonstrations, the damping must be predetermined by using the method proposed in \cite{ras}. However, in real experiments, the demonstrations can provide state information but not damping, meaning that both stiffness and damping are unknown before estimating the stiffness. 

To handle this problem, a modified method is proposed in this paper that can estimate both stiffness and damping from the demonstrations. The method inherits the effectively of \cite {ras} while also utilizing the prior knowledge that $\boldsymbol{K}_{t}$ is a symmetric matrix, resulting in a more reliable stiffness estimation from demonstrations. 

The most commonly available variable impedance control assumes that the damping is constant or corresponds to the stiffness matrices, thereby maintaining critical damping of the system. Then the damping in demonstrations can be also assumed to be constant or correspond to stiffness.  

Firstly, assume that the damping is constant and specified manually, $\boldsymbol{\hat{B}}_{t}$ for equation (\ref{equ:5}) can be calculated as:
\begin{equation}\label{equ:6}
	\boldsymbol{\hat{B}}_{t} = \tilde{\boldsymbol{y}}_{t}\tilde{\boldsymbol{x}}^{\mathrm{T}}_{t}(\tilde{\boldsymbol{x}}_{t}\tilde{\boldsymbol{x}}^{\mathrm{T}}_{t})^{-1}, 	
\end{equation}
or
\begin{equation}\label{equ:7}
	\boldsymbol{\hat{B}}_{t} = \tilde{\boldsymbol{y}}_{t}(\tilde{\boldsymbol{x}}^{\mathrm{T}}_{t}\tilde{\boldsymbol{x}}_{t})^{-1}\tilde{\boldsymbol{x}}^{\mathrm{T}}_{t}.	
\end{equation}
Taking equation (\ref{equ:6}) for example, let $\boldsymbol{T}_{t}$ represents $\tilde{\boldsymbol{y}}_{t}\tilde{\boldsymbol{x}}_{t}^{\mathrm{T}}$, 
The results of (\ref{equ:6}) can be reformulated as 
\begin{equation}\label{equ:8}
	B_{t_{1}}||B_{t_{2}}||\cdots||B_{t_{N^2}} =T_{t_{1}}||T_{t_{2}}||\cdots||T_{t_{N^2}}\boldsymbol{P},
\end{equation}
where the symbol $||$ denotes concatenation, and for  $i=1, 2, \cdots, N^2$ with $\boldsymbol{B}_{t}$ and $\boldsymbol{T}_{t}$ denote for the i-th element in $\boldsymbol{B}_{t_{i}}$ and $\boldsymbol{T}_{t_{i}}$, respectively. The index in the matrix increases from left to right and top to bottom. 
\begin{equation}\label{equ:9}
	\boldsymbol{P}=\begin{bmatrix}
		(\tilde{\boldsymbol{x}}_{t}\tilde{\boldsymbol{x}}_{t}^{\mathrm{T}})^{-1}   & \boldsymbol{0} & \cdots  & \boldsymbol{0}\\ 
		\boldsymbol{0} &(\tilde{\boldsymbol{x}}_{t}\tilde{\boldsymbol{x}}_{t}^{\mathrm{T}})^{-1}  & \cdots & \boldsymbol{0}\\ 
		\vdots & \vdots &\ddots   &\vdots  \\ 
		\boldsymbol{0} &  \cdots & \boldsymbol{0} & (\tilde{\boldsymbol{x}}_{t}\tilde{\boldsymbol{x}}_{t}^{\mathrm{T}})^{-1} 
	\end{bmatrix},
\end{equation}
where $(\tilde{\boldsymbol{x}}_{t}\tilde{\boldsymbol{x}}_{t}^{\mathrm{T}})^{-1}$ is a $N\times N$ matrix while $\boldsymbol{P}$ is a $N^2\times N^2$ matrix. Therefore, directly using the least squares method
is equivalent to first use $\tilde{\boldsymbol{x}}_{t}$ in equation (\ref{equ:5}) times $N\times N$ $N\times N $ matrices as:
$$\begin{bmatrix}
	1   & 0 & \cdots  & 0\\ 
	0 &0  & \cdots & 0\\ 
	\vdots & \vdots &\ddots    &\vdots   \\ 
	0 &  \cdots & 0 & 0 
\end{bmatrix},\begin{bmatrix}
	0   & 1 & \cdots  & 0\\ 
	0 &0  & \cdots & 0\\ 
	\vdots & \vdots &\ddots    &\vdots   \\ 
	0 &  \cdots & 0 & 0 
\end{bmatrix}, \cdots, \begin{bmatrix}
	0   & 0 & \cdots  & 0\\ 
	0 &0  & \cdots & 0\\ 
	\vdots & \vdots &\ddots    &\vdots   \\ 
	0 &  \cdots & 0 & 1 
\end{bmatrix},$$
to acquire $N\times N$ vectors, then use the least squares method to calculate the weights for each calculated vectors, which can recombine into the $\boldsymbol{B}_{t}$.
However, the $N\times N$ matrices are not symmetry, which means that the recombined matrix is unlikely to be symmetric. In this paper the modified method is proposed to change these matrices, whereby $\tilde{\boldsymbol{x}}_{t}$ in equation (\ref{equ:5}) first times $N(N+1)/2$ $N\times N $ symmetry matrices that are denoted as $\boldsymbol{M}_{1}$ to $\boldsymbol{M}_{N(N+1)/2}$. The matrices are constructed as follows:
$$\begin{bmatrix}
	1   & 0 & \cdots  & 0\\ 
	0 &0  & \cdots & 0\\ 
	\vdots & \vdots &\ddots    &\vdots   \\ 
	0 &  \cdots & 0 & 0 
\end{bmatrix},\begin{bmatrix}
	0   & 1 & \cdots  & 0\\ 
	1 &0  & \cdots & 0\\ 
	\vdots & \vdots &\ddots    &\vdots   \\ 
	0 &  \cdots & 0 & 0 
\end{bmatrix}, \cdots, \begin{bmatrix}
	0   & 0 & \cdots  & 0\\ 
	0 &0  & \cdots & 0\\ 
	\vdots & \vdots &\ddots    &\vdots   \\ 
	0 &  \cdots & 0 & 1 
\end{bmatrix}.$$
By times these matrices,  $N(N+1)/2$ vectors are acquired. Equation (\ref{equ:5}) can be reformulated as:
\begin{equation}\label{equ:10}
	\sum_{i=1}^{N(N+1)/2} w^t_{i}\boldsymbol{M}_{i}
	\tilde{\boldsymbol{x}}_{t} =\tilde{\boldsymbol{y}}_{t},	
\end{equation} 
where $w^t_{i}$ are the weight to be calculated, where $i$ ranges from $1$ to $N(N+1)/2$. In equation (\ref{equ:10}), only the weights $w^t_{1}, w^t_{2}, ..., w^t_{N(N+1)}$ are unknown and can be estimated by using the least squared. Once determined, they can be combined with $\boldsymbol{M}_{i}$ 
to obtain an approximation of the stiffness as follows:
\begin{equation}\label{equ:11}
	\boldsymbol{B}_{t}=\sum_{i=1}^{N(N+1)/2} w^t_{i}\boldsymbol{M}_{i}. 	
\end{equation}
However,   there remains a possibility that the combined matrices may not be positive definite. Therefore, the nearest-SPD matrix method must still be applied to $\boldsymbol{B}_{t}$ to estimate the stiffness. This process involves decomposing the matrix using polar decomposition and then recombining it as previously described, omitting the first step. The proposed method estimates fewer unknown parameters than direct least squares, which suggests that it has the potential to achieve higher accuracy when estimating stiffness from demonstrations.

Secondly, assume that the damping value is an unknown constant, which is more representative of real experiments. The stiffness is still decomposed using the method described earlier, which involves acquiring $N(N+1)/2$ unknown parameters. Next, all $N(N+1)/2+1$ parameters, including $N(N+1)/2$ unknown stiffness parameters and the unknown damping constant, are all estimated using the least squares method. Subsequently, the estimated stiffness is acquired by recombining matrices with the calculated parameters and applying the nearest-SPD method. 

Thirdly, assume that the damping is unknown but varies in correspondence with the stiffness, which means that assuming the demonstrations are critical damped. In other words, the stiffness and damping matrices are SPD and satisfy that: 
\begin{equation}\label{equ:12}
	\begin{cases}
		\boldsymbol{K}_{t}=\boldsymbol{T}_{t}^{\mathrm{T}} \boldsymbol{\Lambda_{t}}\boldsymbol{T}_{t},\\
		\boldsymbol{D}_{t}=\boldsymbol{T}_{t}^{\mathrm{T}} (\zeta\boldsymbol{\Lambda^{\frac{1}{2}}_{t}})\boldsymbol{T}_{t},  \\
	\end{cases}
\end{equation}
where $\boldsymbol{T}_{t}$ and $\boldsymbol{\Lambda}$ are the eigenvector and eigenvalue of the $\boldsymbol{K}_{t}$, and $\zeta \in \mathbb{R}^{+}$ is a manually designed parameter. 

The approach of estimating the complete stiffness matrix involves  decomposing the stiffness and damping components into the 
$N(N+1)/2$ distinct submatrices, , as detailed earlier, and subsequently estimating the $N(N+1)$
unknown parameters via least squares, frequently yields bad outcomes. This is primarily due to the method's inability to satisfy the inherent constraint relationship between the  stiffness matrix $\boldsymbol{K}_{t}$ and the damping matrix $\boldsymbol{D}_{t}$.
Moreover, the constraint is non-convex, which further complicates the optimization process, leading to the failure of conventional convex optimization techniques.

To address this issue, the evolutionary optimization method presents a compelling solution. It offers a balance between approximate accuracy and computational efficiency. In this paper,  the Covariance Matrix Adaptation Evolution Strategy (CMA-ES) is employed, which is highly competitive in this scenario.

Normally, there is no need to assume that the demonstrations are critically damped, as an unknown constant damping value is frequently used. In addition, during reproduction using the learned stiffness, $\boldsymbol{D}_{t}$ can be chosen by values learned from demonstrations or based on the desired response (e.g., critical damping) of the interaction system (\ref{equ:3}) chosen by experts.

\subsection{Stiffness Learning and Reproduction}\label{sec3_3}

The successful completion of most manipulation tasks relies heavily on the ability to perceive both force and position. When employing learned variable impedance skills, force and position are interdependent. By achieving suitable impedance and position, the ideal force can be attained, allowing for stiffness to be reproduced based on position.

GMMs are highly effective at establishing relationship between different variables, but their performance may be unsatisfactory when only one demonstration is provided \cite{stifftro}. To face the conditions that either one or more demonstrations are provided, 
in this paper, $\boldsymbol{K}_{t}$ is first performed a Cholesky decomposition as $\boldsymbol{K}_{t}=\boldsymbol{L}_{t}^{\mathrm{T}}\boldsymbol{L}_{t}$, where  $\boldsymbol{L}_{t}$ a lower triangular matrix and then the vectorization of $\boldsymbol{L}_{t}$ is modeled as:
\begin{equation}\label{equ:13}
	\tilde{\boldsymbol{L}}_{t} =\Theta (\boldsymbol{s}_{t})^{\mathrm{T}} \boldsymbol{w}.
\end{equation}
where $\Theta(\boldsymbol{s}_{t})$ is a vector of basis functions that depend on $\boldsymbol{s}_{t}$, which is the position $\boldsymbol{x}_{t}$ ,  and $\boldsymbol{w}$ is a weight vector to be learned.
Therefore, after estimating stiffness and calculating the $\boldsymbol{L}_{t}$ from demonstration, the model can be obtained by minimizing the following objective function:
\begin{equation}\label{equ:14}
	J(\boldsymbol{w}) =\frac{1}{2}\sum_{n=1}^{N_{d}} \sum_{t=1}^{t_{m}}\begin{Bmatrix}
		\boldsymbol{w}^\mathrm{T}\Theta (\boldsymbol{s}_{t})_{n}-(\boldsymbol{x}_{t})_{n}
	\end{Bmatrix}^2+\frac{1}{2}\lambda\boldsymbol{w}^\mathrm{T}\boldsymbol{w},
\end{equation}
where $\lambda$ is a positive constant to circumvent over-fitting, $n$ denotes the index of a demonstration, $t_{m}$ is the total number of samples in one demonstration and $N_{d}$ represents the total number of demonstrations. 

The optimal solutions of the objective function can been represented as \cite{regression}:
\begin{equation}\label{equ:15}
	\boldsymbol{w}^{*} =  \Theta(\Theta^{\mathrm{T}}\Theta+\lambda I)^{-1}\boldsymbol{x},
\end{equation}
where $\Theta = [\Theta(\boldsymbol{s}_{1}), \Theta(\boldsymbol{s}_{2}) \cdots, \Theta(\boldsymbol{s}_{N_{d}t_{N}})]$, $I$ is the $N_{d}t_{N}$-dimensional identity matrix and with a bit abuse of notations,  $\boldsymbol{x}$ mean concatenating all the elements of position in demonstrations, respectively. Substitute the optimal $\boldsymbol{w}^{*}$ into (\ref{equ:13}),  the vector $\tilde{\boldsymbol{L}}$ can be calculated as:
\begin{equation}\label{equ:16}
	\tilde{\boldsymbol{L}}_{t} =\Theta(\boldsymbol{s}_{t})^{\mathrm{T}}\Theta(\Theta^{\mathrm{T}}\Theta+\lambda \boldsymbol{I})^{-1}\boldsymbol{x}
\end{equation}

To avoid the explicit definition of the basis functions in (\ref{equ:16}), the kernel trick is used to express the inner product as follows:
\begin{equation}\label{equ:17}
	\begin{split}
		\boldsymbol{Ker}= \qquad \qquad \qquad \qquad \qquad \qquad \qquad 
		\qquad \qquad \qquad
		\qquad  \\ \begin{bmatrix}
			ker(\boldsymbol{s}_{1},\boldsymbol{s}_{1}) & ker(\boldsymbol{s}_{1},\boldsymbol{s}_{2}) & \cdots & ker(\boldsymbol{s}_{1},\boldsymbol{s}_{N_{d}t_{m}})\\
			ker(\boldsymbol{s}_{2},\boldsymbol{s}_{1}) &ker(\boldsymbol{s}_{2},\boldsymbol{s}_{2})  & \cdots   & ker(\boldsymbol{s}_{2},\boldsymbol{s}_{N_{d}t_{m}})\\
			\vdots  & \vdots & \ddots  & \vdots\\
			ker(\boldsymbol{s}_{N_{d}t_{m}},\boldsymbol{s}_{1}) & ker(\boldsymbol{s}_{N_{d}t_{m}},\boldsymbol{s}_{2}) & \cdots  & ker(\boldsymbol{s}_{N_{d}t_{m}},\boldsymbol{s}_{N_{d}t_{m}})
		\end{bmatrix}
	\end{split}	
\end{equation}
where $ker(\boldsymbol{s}_{i},\boldsymbol{s}_{j})=\Theta(\boldsymbol{s}_{i})^{\mathrm{T}}\Theta(\boldsymbol{s}_{j}) \quad   \forall i,j \in [1,2 \cdots, {N_{d}t_{m}}]$.

Thus, equation (\ref{equ:16}) can be reformulated as follows:
\begin{equation}\label{equ:18}
	\tilde{\boldsymbol{L}} =\boldsymbol{ker}^{*}(\boldsymbol{Ker}+\lambda \boldsymbol{I})^{-1}\boldsymbol{x},
\end{equation}
where $\boldsymbol{ker}^{*} = [ker(\boldsymbol{s},\boldsymbol{s}_{1}),ker(\boldsymbol{s},\boldsymbol{s}_{2})\cdots,ker(\boldsymbol{s},\boldsymbol{s}_{N_{d}t_{m}})].$

The kernel function utilizes the Gaussian kernel, defined as follows:
\begin{equation}\label{equ:19}
	ker(\boldsymbol{s}_{i},\boldsymbol{s}_{j})=e^{(-h \left \|(\boldsymbol{s}_{i}-\boldsymbol{s}_{j})  \right \|^2)},
\end{equation}
where $h$ is a manually designed constant.

Therefore, when acquiring the position and interaction force, the vectorization of $\tilde{\boldsymbol{L}}$ is calculated from (\ref{equ:18}), and the lower triangular matrix  $\boldsymbol{L}$ can be recombination from the vectorization of $\tilde{\boldsymbol{L}}$, afterwards, the stiffness is calculated as $\boldsymbol{K}=\boldsymbol{L}^{\mathrm{T}}\boldsymbol{L}$ that satisfies the SPD constraints.  

The purpose of reproducing the stiffness matrix is to shape the compliance of the robot like that of the human teacher.
The control torque $\boldsymbol{f}^{c}$ in (\ref{equ:1}) for the robot can be calculated using the reproduced stiffness and a chosen $\boldsymbol{K}^{v}$. This torque is then transformed to joint torques $\boldsymbol{\tau}$ of the robot using the Jacobian transpose $J^{\mathrm{T}}$ as follows:
$\boldsymbol{\tau}=J^{\mathrm{T}} \boldsymbol{f}^{c}$. However, directly applying the learned stiffness for interaction tasks can potentially lead to instability or unsafe behavior in the system. To address this issue, the next section introduces a novel Lyapunov function that provides a straightforward implementation condition to ensure the system's passivity.

\subsection{Novel Lyapunov Functions}
The Lyapunov-based approach is founded on a robust energy-based concept. To ensure stability, the system's potential energy is dissipated.
For the utilization of learned stiffness, ensuring the passivity of safe robot operation is essential.

Firstly, considering a novel Lyapunov function with $\alpha$ being a  constant:
\begin{equation}\label{equ:20}
	V(\boldsymbol{e},\dot{\boldsymbol{e}},t)=\frac{(\dot{\boldsymbol{e}}+\alpha \boldsymbol{e})^{\mathrm{T}} \boldsymbol{H} (\dot{\boldsymbol{e}}+\alpha \boldsymbol{e})}{2}+\frac{\boldsymbol{e}^{\mathrm{T}} \boldsymbol{K}_t\boldsymbol{e}}{2},
\end{equation}
whose differentiating can be calculated as:
\begin{equation}\label{equ:21}
	\begin{aligned}
			\dot{V}(\boldsymbol{e},\dot{\boldsymbol{e}},t)=(\dot{\boldsymbol{e}}+\alpha \boldsymbol{e})^{\mathrm{T}}\boldsymbol{F}_{ext}-\alpha\boldsymbol{e}^\mathrm{T}\boldsymbol{K}_ {t}\boldsymbol{e}-\boldsymbol{\dot{e}}^\mathrm{T}\boldsymbol{D}_ {t}\boldsymbol{\dot{e}} \\  +\frac{1}{2} \boldsymbol{e}^\mathrm{T} \boldsymbol{\dot{K}}_{t}\boldsymbol{e}+\alpha \boldsymbol{\dot{e}}^\mathrm{T}\boldsymbol{H}\boldsymbol{\dot{e}}
			+\boldsymbol{\dot{e}}^\mathrm{T}(\alpha^2\boldsymbol{H}-\alpha \boldsymbol{D}_{t})\boldsymbol{e}\\
			=(\dot{\boldsymbol{e}}+\alpha \boldsymbol{e})^{\mathrm{T}}\boldsymbol{F}_{ext}+(\boldsymbol{\dot{e}}+\frac{1}{2} \alpha \boldsymbol{e})^{\mathrm{T}}(\alpha\boldsymbol{H}-\boldsymbol{D}_{t})(\boldsymbol{\dot{e}}+\frac{1}{2} \alpha \boldsymbol{e})
			\\
			\quad+\boldsymbol{e}^\mathrm{T}(\frac{1}{2} \boldsymbol{\dot{K}}_ {t}-\alpha \boldsymbol{K}_ {t}-\frac{1}{4} \alpha^2 (\alpha\boldsymbol{H}-\boldsymbol{D}_{t}))\boldsymbol{e}.
	\end{aligned}	
\end{equation}

Therefore, if the matrices $\alpha^2\boldsymbol{H}-\alpha \boldsymbol{D}_{t}$ and $\frac{1}{2} \boldsymbol{\dot{K}}_ {t}-\alpha \boldsymbol{K}_ {t}-\frac{1}{4} \alpha^2 (\alpha\boldsymbol{H}-\boldsymbol{D}_{t})$ are negative semidefinite,  the following inequalities always hold:
\begin{equation}\label{equ:26}
	V(t)-V(0) \le \int_{0}^{t} (\dot{\boldsymbol{e}}+\alpha \boldsymbol{e})^{\mathrm{T}}\boldsymbol{F}_{ext}dt,		
\end{equation}
and system is passivity.  With the Lyapunov function introduced in this paper, the finding contrasts with the stability conditions outlined in \cite{stabilitytro} as $\boldsymbol{\dot{D}}_{t}$ is deemed unnecessary for consideration and has no influence on the system's passivity.

Considering the following one dimensional scalar system:
\begin{equation}\label{equ:27}
	\ddot{q}+d(t)\dot{q}+k(t)q = 0
\end{equation}
where critical damping is chosen as $d(t)=2\varsigma \sqrt{k(t)} $, with $\varsigma$ being a positive constant.
Using the stability conditions proposed in \cite{stabilitytro}, the constraints are:
\begin{equation}\label{equ:28}
	\begin{cases}
		&  2\varsigma\sqrt{k(t)}<\alpha \\
		&  \dot{k}(t)<\frac{2\alpha \sqrt{k(t)}^3}{\sqrt{k(t)}+2\varsigma\alpha} 
	\end{cases}  
\end{equation}
Under the proposed stability conditions, the constraint becomes:
\begin{equation}\label{equ:29}
	\begin{cases}
		&  2\varsigma\sqrt{k(t)}<\alpha \\
		&  \dot{k}(t)<2\alpha k(t)+\frac{1}{2}\alpha^2(\alpha-2\varsigma \sqrt{k(t)}) 
	\end{cases}
\end{equation}
Compare $2\alpha k(t)+\frac{1}{2}\alpha^2(\alpha-2\varsigma \sqrt{k(t)})$ with $\frac{2\alpha \sqrt{k(t)}^3}{\sqrt{k(t)}+2\varsigma\alpha}$ is equivalent to
compare $6\varsigma k(t)+\alpha \sqrt{k(t)}+2\alpha^2\varsigma$ with $4\alpha \varsigma^2\sqrt{k(t)}$. Given the first condition, it can be seen that $4\alpha \varsigma^2\sqrt{k(t)}<2\alpha^2 \varsigma $  (\ref{equ:28}), while $6\varsigma k(t)+\alpha \sqrt{k(t)}+2\alpha^2\varsigma>2\alpha^2 \varsigma$. Therefore, the proposed conditions are strictly less restrictive than those  proposed in \cite{stabilitytro} when considering one dimensional critical damping system.

Moreover, another Lyapunov function is introduced:

\begin{equation}\label{equ:22}
	V_2(\boldsymbol{e},\dot{\boldsymbol{e}},t)=\frac{(\dot{\boldsymbol{e}}+\alpha \boldsymbol{H}^{-1} \boldsymbol{e})^{\mathrm{T}} \boldsymbol{H} (\dot{\boldsymbol{e}}+\alpha \boldsymbol{H}^{-1} \boldsymbol{e})}{2}+\frac{\boldsymbol{e}^{\mathrm{T}} \boldsymbol{\beta }_{t} \boldsymbol{e}}{2},
\end{equation}
where $\alpha$ is a constant, and $\boldsymbol{\beta }$ is positive definite matrix. It can be easily proved that both $\boldsymbol{e}$ and $\boldsymbol{\dot{e}}$ are zeros when $	V_2(\boldsymbol{e},\dot{\boldsymbol{e}},t)$ is zeros. Then the differentiating of the function can be calculated as:
\begin{equation}\label{equ:25}
	\begin{split}
		\dot{V}_2(\boldsymbol{e},\dot{\boldsymbol{e}},t)=(\dot{\boldsymbol{e}}+\alpha \boldsymbol{H}^{-1} \boldsymbol{e})^{\mathrm{T}}\boldsymbol{F}_{ext}+ \boldsymbol{e}^\mathrm{T}(\boldsymbol{\frac{1}{2} \dot{\beta}_{t}-\alpha\boldsymbol{H}^{-1}\boldsymbol{K}_ {t} })\boldsymbol{e} \\ +\boldsymbol{\dot{e}}^\mathrm{T}(\alpha-\boldsymbol{D}_ {t})\boldsymbol{\dot{e}}  +  \boldsymbol{\dot{e}}^\mathrm{T}(\alpha^2\boldsymbol{H}^{-1}-\boldsymbol{K}_{t}-\alpha \boldsymbol{H}^{-1}\boldsymbol{D}_{t}+\boldsymbol{\beta }_{t})\boldsymbol{\dot{e}}
	\end{split}
\end{equation}
Therefore, by setting $\boldsymbol{\beta }_{t}=\boldsymbol{K}_{t}+\alpha \boldsymbol{H}^{-1}\boldsymbol{D}_{t}-\alpha^2\boldsymbol{H}^{-1}$ and ensuring that $(\alpha-\boldsymbol{D}_{t})$, and $(\boldsymbol{\dot{K}}_{t}+\alpha\boldsymbol{H}^{-1}\boldsymbol{\dot{D}}_{t}-2\alpha\boldsymbol{H}^{-1}\boldsymbol{\dot{K}}_{t}$ are negative semidefinite, $\boldsymbol{\beta }$ must be positive definite, and the system demonstrates passivity. For one-dimensional variable control, the function is identical to the one proposed in \cite{stabilitytro}. 

The stability conditions presented in \cite{stabilitytro} can be reformulated  as:
\begin{equation}\label{equ:36}
	\boldsymbol{\dot{K}}_{t}+\alpha\boldsymbol{\dot{D}}_{t}-2\alpha\boldsymbol{K}_{t} \preceq \boldsymbol{0}, \ \ \ s.t.  \alpha<=\underset{t}{\min}{\frac{\underline{\lambda}(\boldsymbol{D}_t)}{\overline{\lambda} (\boldsymbol{H})}} 	
\end{equation}
In contrast, the proposed stability conditions can be expressed as:
\begin{equation}\label{equ:37}
	\boldsymbol{\dot{K}}_{t}+\alpha\boldsymbol{H}^{-1}\boldsymbol{\dot{D}}_{t}-2\alpha\boldsymbol{H}^{-1}\boldsymbol{K}_{t} \preceq \boldsymbol{0},\ \ \ s.t.  \alpha<=\underset{t}{\min}{\underline{\lambda}(\boldsymbol{D}_t)}, 	
\end{equation}
where  $\underline{\lambda}$ and $\overline{\lambda}$ denote the smallest and largest eigenvalue, respectively. To maintain consistency with the proposed stability condition, the conditions in \cite{stabilitytro} can be rewritten as:
\begin{equation}\label{equ:38}
	\boldsymbol{\dot{K}}_{t}+\alpha\boldsymbol{\dot{D}}_{t}/	\overline{\lambda} (\boldsymbol{H})-2\alpha\boldsymbol{K}_{t}/	\overline{\lambda} (\boldsymbol{H}) \preceq \boldsymbol{0},\ \ \ s.t.  \alpha<=\underset{t}{\min}{\underline{\lambda}(\boldsymbol{D}_t)}, 	
\end{equation}	
Therefore when considering  multi-dimensional variable control, the proposed function may exhibit superior performance. 
\section{Experiment Results and Discussions}\label{sec4}
To evaluate the effectiveness of the proposed algorithm, both simulations and real experiments are conducted. The simulation is implemented in MATLAB and the experiment is implemented using Franka Emika robot in C++ 14 on the platform of CPU Inter Core i5-8300H, Ubuntu 20.04.
The experiments used in the evaluation are summarized as follows:
\begin{itemize}
	
	\item Compare the performance of the stiffness estimating between the proposed algorithm, the method outlined in \cite{ras}, and the convex optimization introduced in \cite{stifftro}.  	
	\item The effectiveness of the proposed approach is evaluated by using the Franka robot to perform a massage task with the learned stiffness.
	
\end{itemize}

\subsection{Simulation of Estimating Stiffness}

\begin{figure}[htbp]
	
	\includegraphics[scale=0.2]{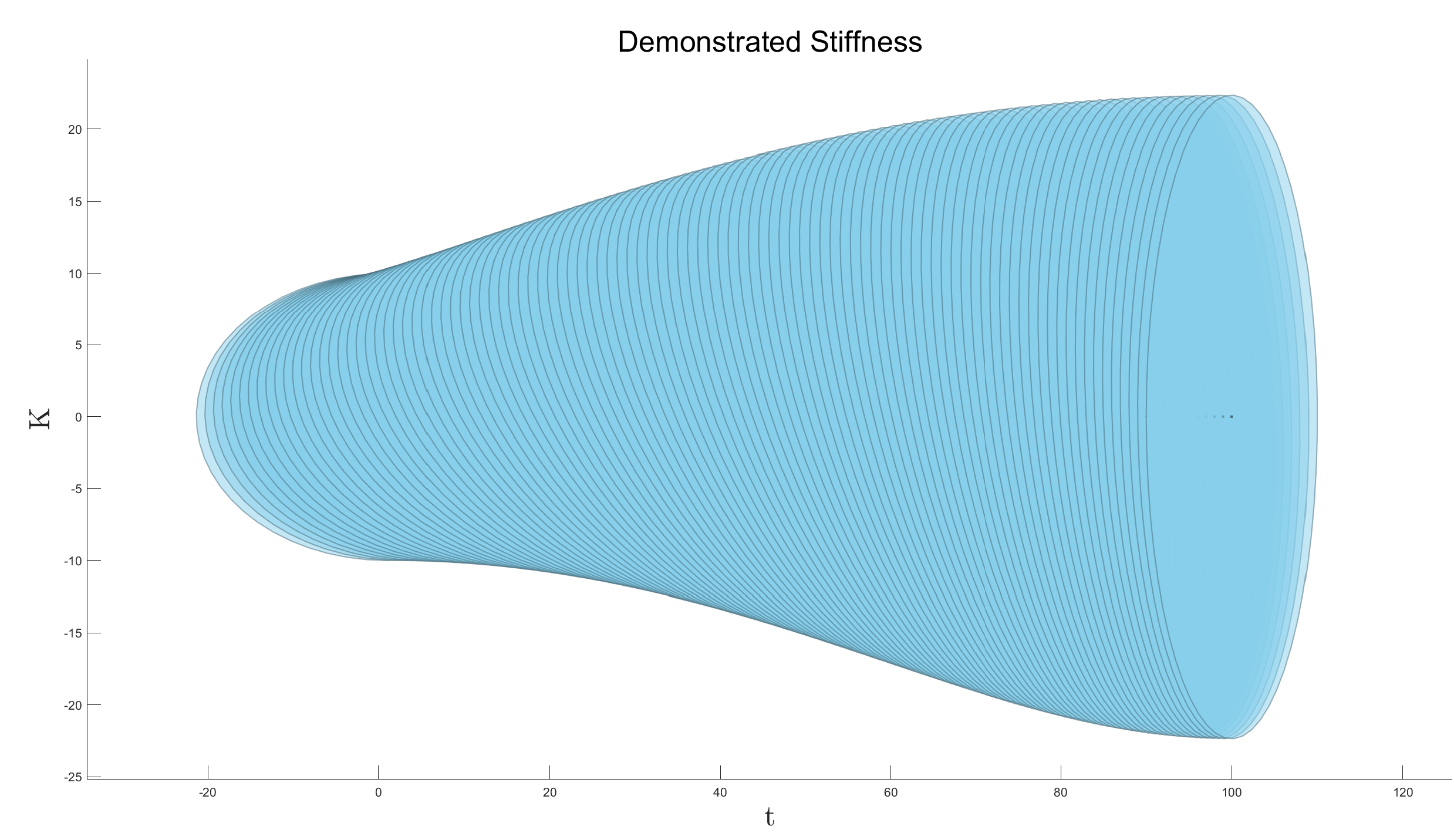} 
	\caption{The demonstrated stiffness ellipsoids are visualized at different time steps.}
	\label{fig3}
\end{figure}

In this section, a self-made dataset is first generated by a 2-degree-of-freedom mass-spring-damper with a manually designed $\boldsymbol{f}_{t}^{e}$, $\boldsymbol{K}_{t}^{p}$ and $\boldsymbol{K}_{t}^{v}$, respectively. The variation of $\boldsymbol{K}_{t}^{p}$ follows the same rule, while  $\boldsymbol{K}_{t}^{v}$ remains a constant matrix, and $\boldsymbol{f}_{t}^{e}$ varies according to a different rule in each demonstration. Then the manually specified stiffness ellipsoids were initialized from a horizontally oriented ellipsoid, which was then continuously rotated clockwise by $\boldsymbol{R}^{\mathrm{T}}\boldsymbol{K}^{p}\boldsymbol{R}$ until it reached a 45-degree rotation, is considered as the ground truth for comparison. The stiffness matrices are shown Fig. \ref{fig3} as stiffness ellipses which were proposed in \cite{bhogan} for graphical visualization of stiffness matrix.

The simulation is performed to estimate the stiffness with a sliding window length $L=3$ for ten different trajectories. Since the ground truth remains unchanged, the estimated stiffness of each trajectory is compared to the ground truth and ten estimation errors can be calculated. The metric errors between different SPD matrices are calculated using: (1) the Affine-invariant, (2) the Log-Euclidean, and (3) the Log-determinant \cite{metric}.                                                                                                                                                                                                                                                                                                                                                                  

First assume that the damping coefficient is known and a constant as $\boldsymbol{D}_{t}=50$. The comparisons between the proposed algorithm and the method outlined in \cite{ras}, as well as the convex optimization introduced in \cite{stifftro} are presented and analyzed. The stiffness in equation \ref{equ:4} is estimated by solving the following convex optimization problem.
\begin{equation}\label{equ:40}
	\mathrm{minimize}\;\;||	\boldsymbol{K}\boldsymbol{X}-\boldsymbol{Y}||_{2},  \; \;  \mathrm{subject \; to}\; \boldsymbol{K} \succeq \boldsymbol{0},
\end{equation}
where $\boldsymbol{K} \succeq \boldsymbol{0}$ denotes that $\boldsymbol{K}$ is a positive semi-definite matrix.

\begin{figure}[htbp]
	
	\includegraphics[scale=0.2]{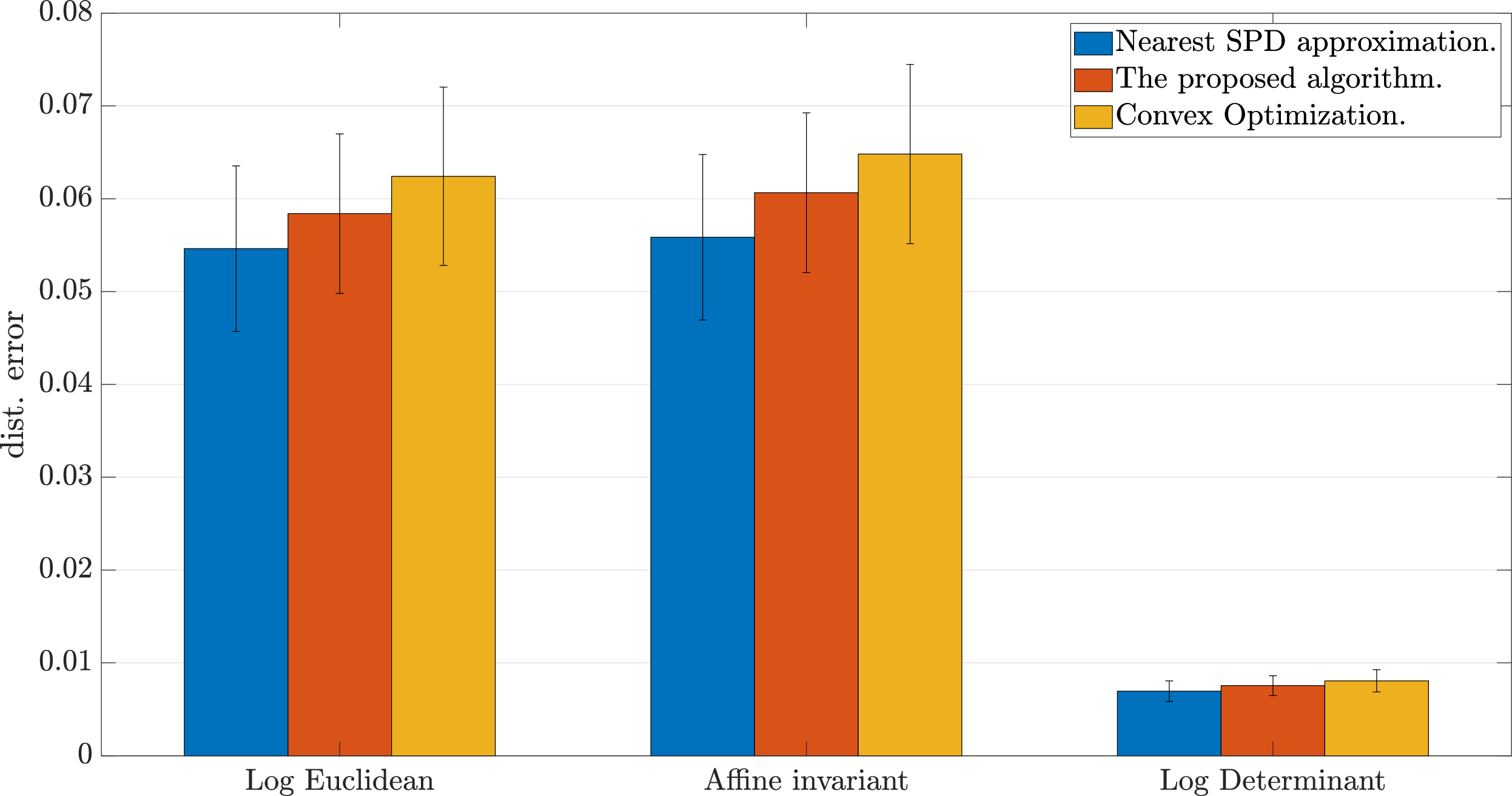} 
	\caption{Considering a known constant damping, compare the performance of the estimated stiffness using the proposed algorithm with nearest-SPD approximation and the convex optimization.}
	\label{fig4}
\end{figure}

The average computation time for the nearest–SPD approximation \cite{ras}, convex optimization \cite{stifftro}, and the proposed algorithm are 0.23, 312, and 0.24 ms per time step, respectively. The results of these algorithms compared with the ground truth are shown in Fig. \ref{fig4}.

It can be seen from Fig. \ref{fig4} that all the three approaches can acquire accurate estimates of the stiffness.  
The nearest–SPD approximation \cite{ras} produces the lowest error across all three metrics, indicating that it exhibits slightly better performance in estimating stiffness compare with the ground truth.  However, it is worth noting that the estimation process involves the use of sliding windows in the simulation, and as the actual stiffness varies within these windows, the ground truth may not be identical to the true value but rather a close approximation. The results of the proposed algorithm and convex optimization may be more influenced by varying stiffness, leading to slightly worse performance.

To get a better consistency with the process of the least squares, changing the original optimization problem to the following formulation when using the convex optimization:
\begin{equation}\label{equ:41}
	\mathrm{minimize}\;\;||	\boldsymbol{K}\boldsymbol{X}\boldsymbol{X}^{\mathrm{T}}-\boldsymbol{Y}\boldsymbol{X}^{\mathrm{T}}||_{2},  \; \;  \mathrm{subject \; to}\; \boldsymbol{K} \succeq \boldsymbol{0},
\end{equation}
Then an interesting result is obtained: the estimated stiffness from the proposed algorithm closely aligns with that obtained using the convex optimization method in this cases, demonstrating the effectiveness of the proposed algorithm.
\begin{figure}[htbp]
	
	\includegraphics[scale=0.2]{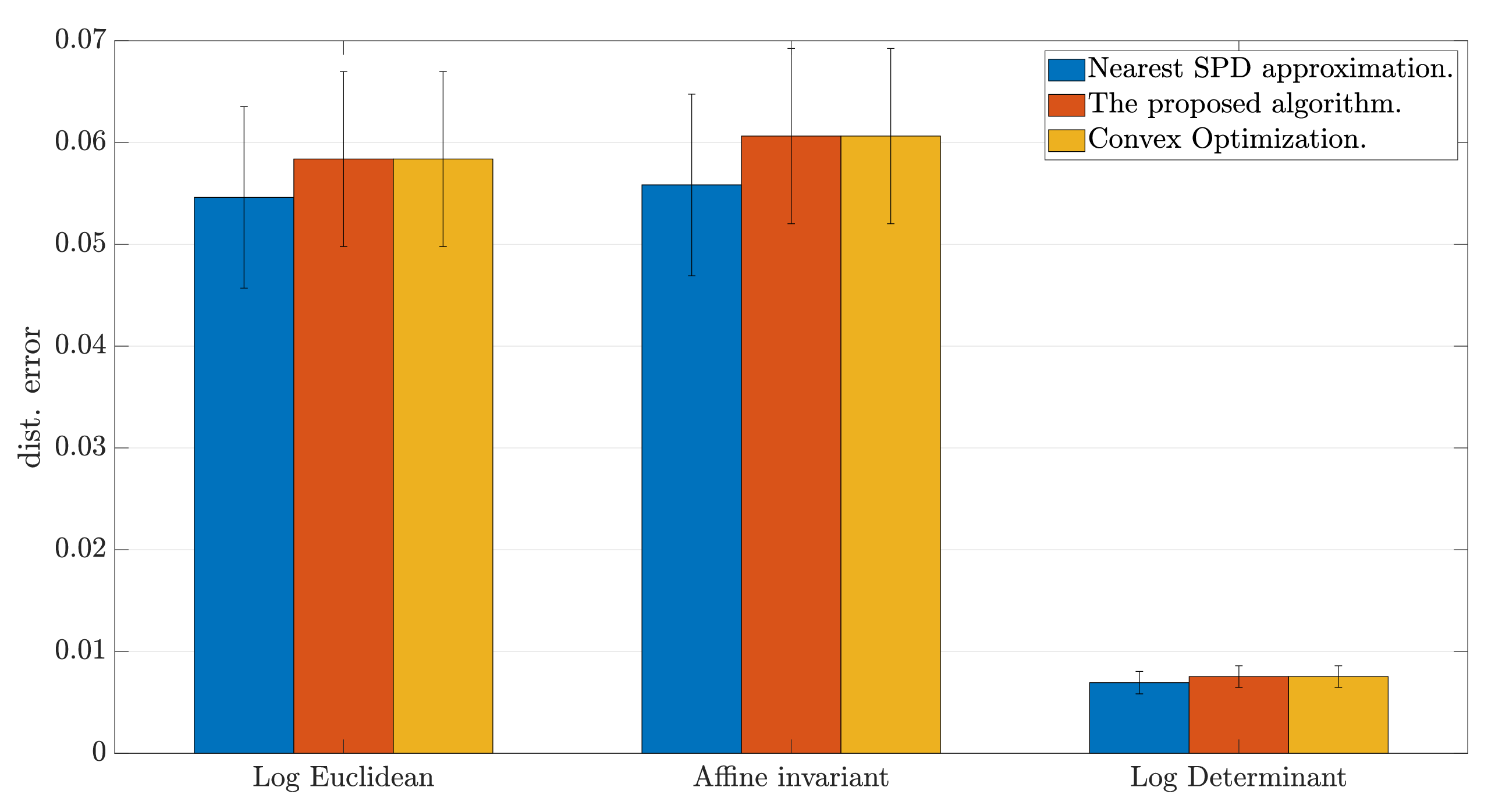} 
	\caption{With known constant damping, compare the performance of the estimated stiffness using the proposed algorithm with nearest-SPD approximation to that using convex optimization to solve the modified objective function.}
	\label{fig5}
\end{figure}

\begin{figure}[htbp]	
	\includegraphics[scale=0.20]{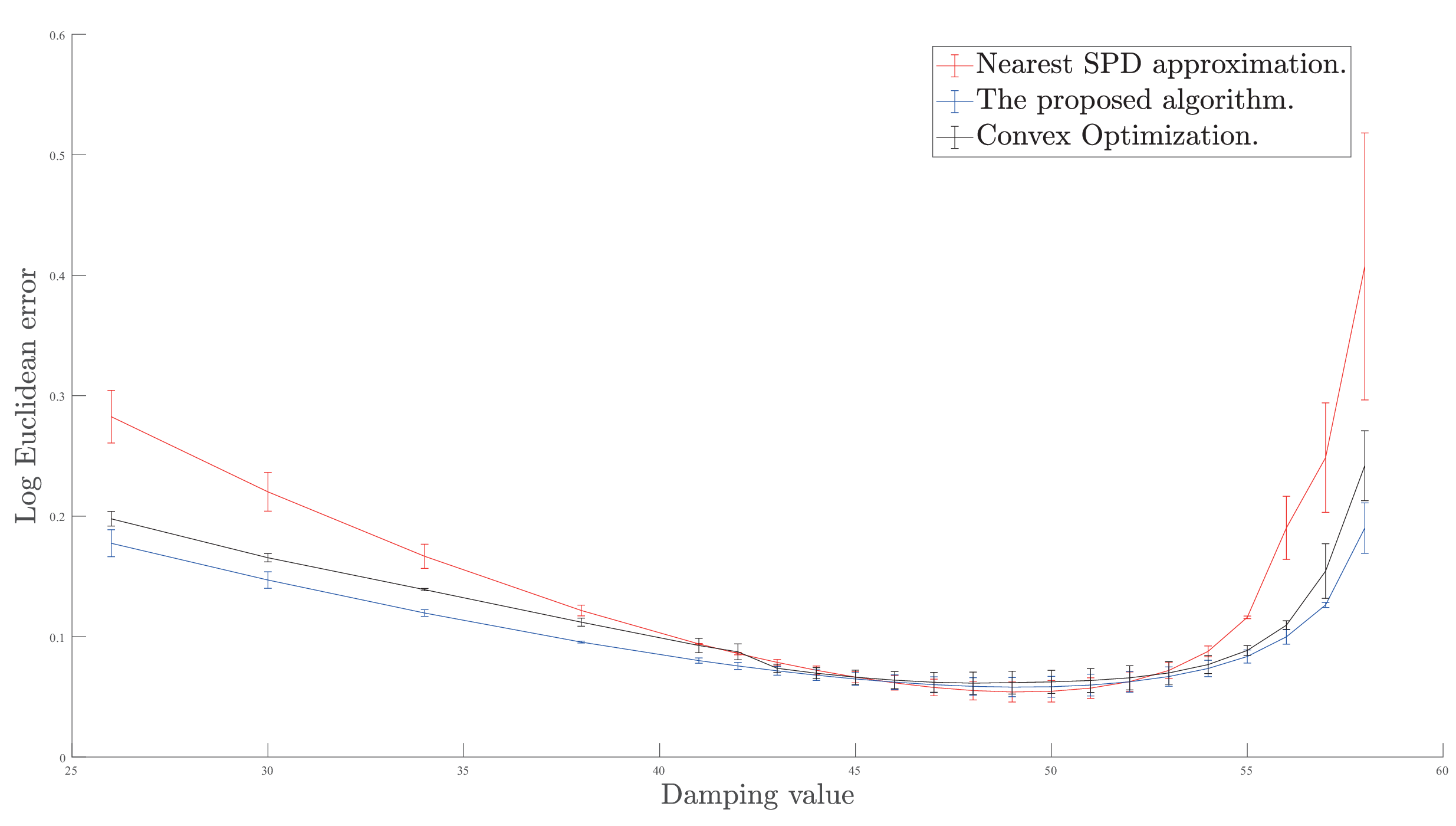}
	\hspace{-2mm}	
	\caption{The stiffness estimation results obtained from applying the proposed algorithm with nearest-SPD approximation and the convex optimization method while varying the damping. }
	\label{fig6}
\end{figure}

Afterwards, a simulation assumes that the damping coefficient is not precisely known but guess it from $26$ to $58$ which are close to the actual value. The performances of estimating stiffness using the three algorithms compare the ground truth are shown in Fig. \ref{fig6}.
Which illustrates the performance of the Log-Euclidean error metric, while the other metrics are not shown as they behave similarly when the damping value is altered. 
The findings indicate that all the approaches are capable of accurately estimating the stiffness, even when different damping values are utilized. However, there are notable distinctions in their performance.  The proposed algorithm exhibits greater robustness in estimating the desired stiffness across a range of damping values.  In contrast, the nearest-SPD approximation and the convex optimization method display a slightly larger variation in stiffness estimates as the damping changes. Specifically, increasing damping error results in decreased accuracy of estimated stiffness for all these three methods. However, the proposed algorithm shows a slower rate of decrease compare with the other two algorithms making it a more suitable choice for applications where damping is unknown constant. 
\begin{figure}[htbp]
	\includegraphics[scale=0.20]{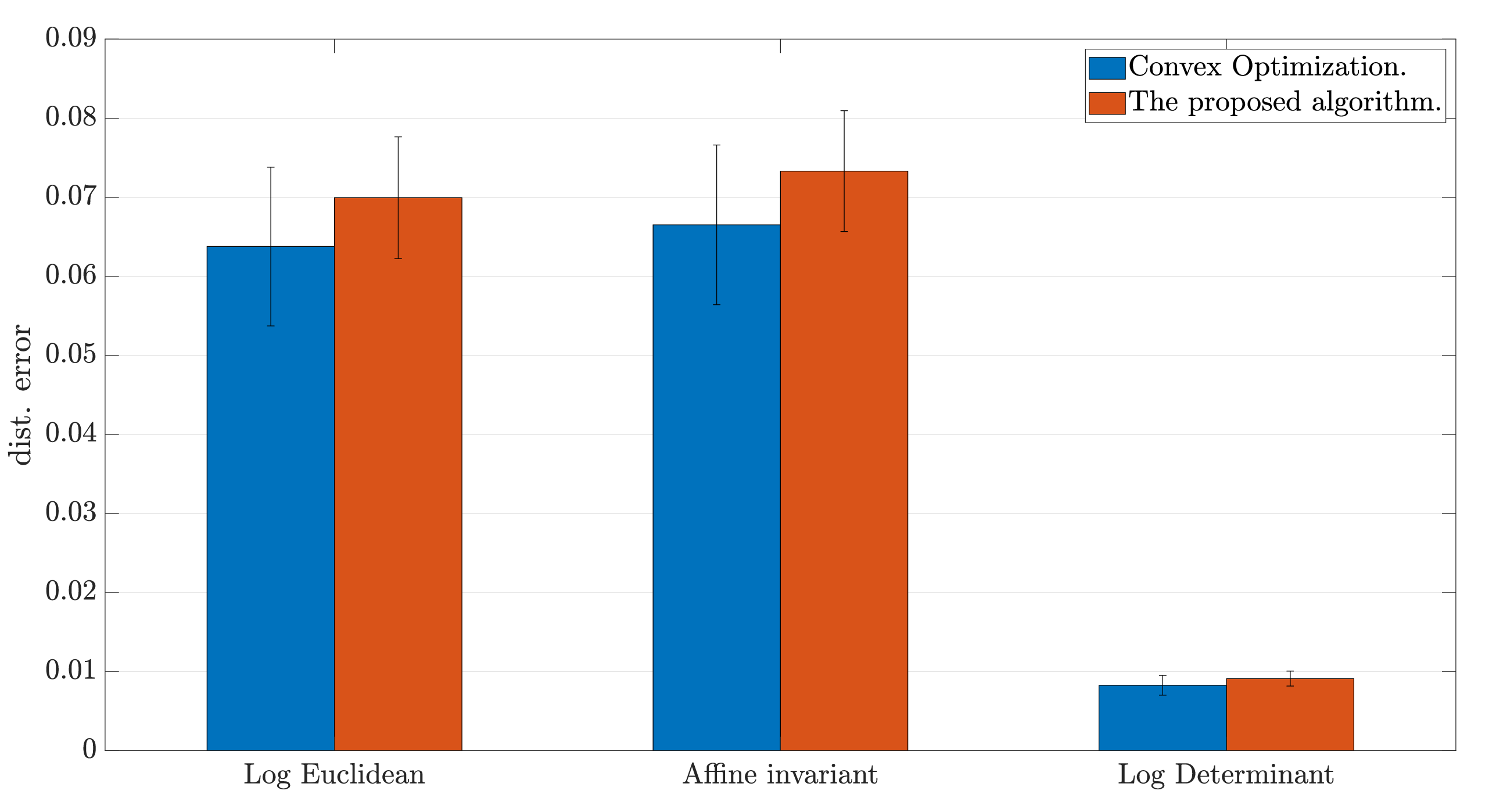} 
	\caption{Compare the performance of the estimated stiffness using the proposed algorithm with the convex optimization when assuming that damping is an unknown constant.}
	\label{fig7}
\end{figure}

\begin{figure}[htbp]
	\includegraphics[scale=0.50]{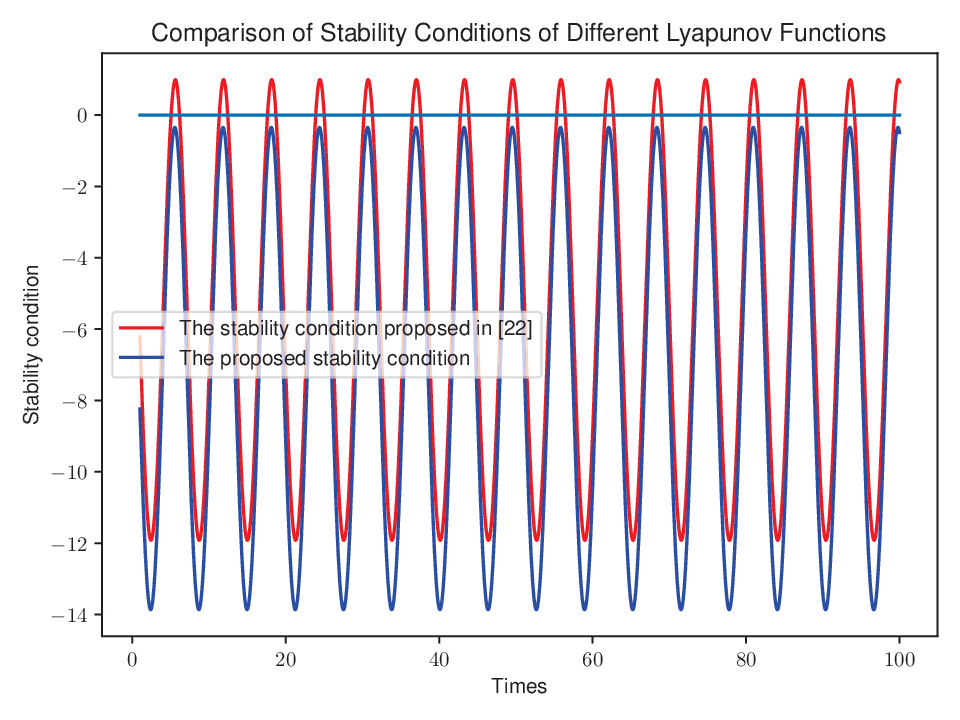} 
	\caption{Compare the stability condition of the proposed Lyapunov function with the condition presented in \cite{stabilitytro}.}
	\label{fig8}
\end{figure}
\begin{figure}[htbp]
	\includegraphics[scale=0.50]{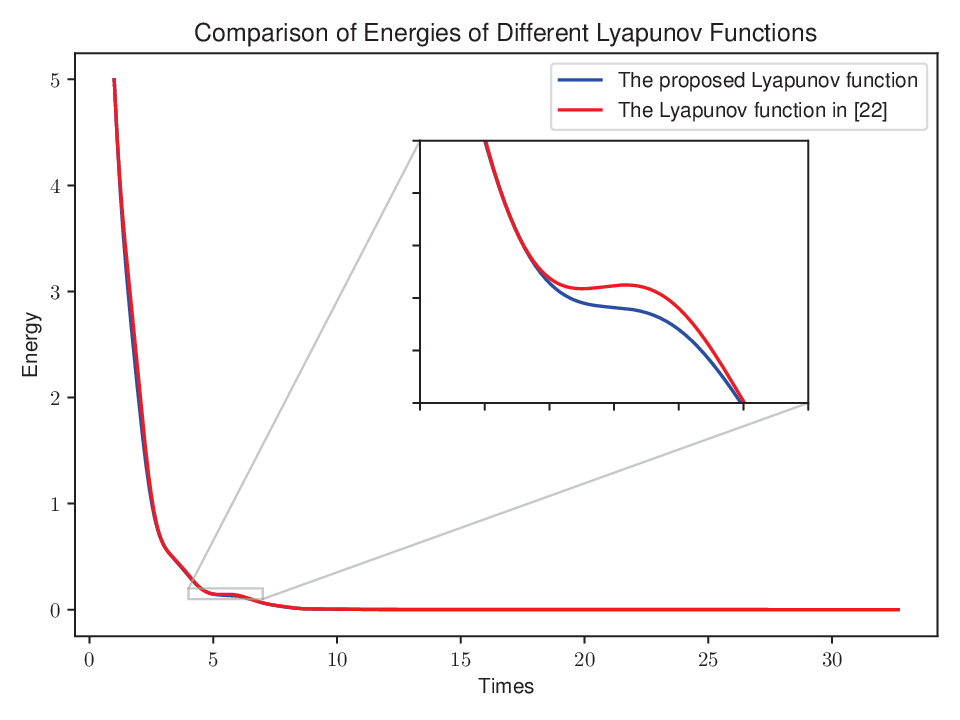} 
	\caption{Compare the energy variation of the proposed Lyapunov function with the performance described in \cite{stabilitytro}.}
	\label{fig9}
\end{figure}
\begin{figure*}[htbp]
	\includegraphics[width=1\textwidth]{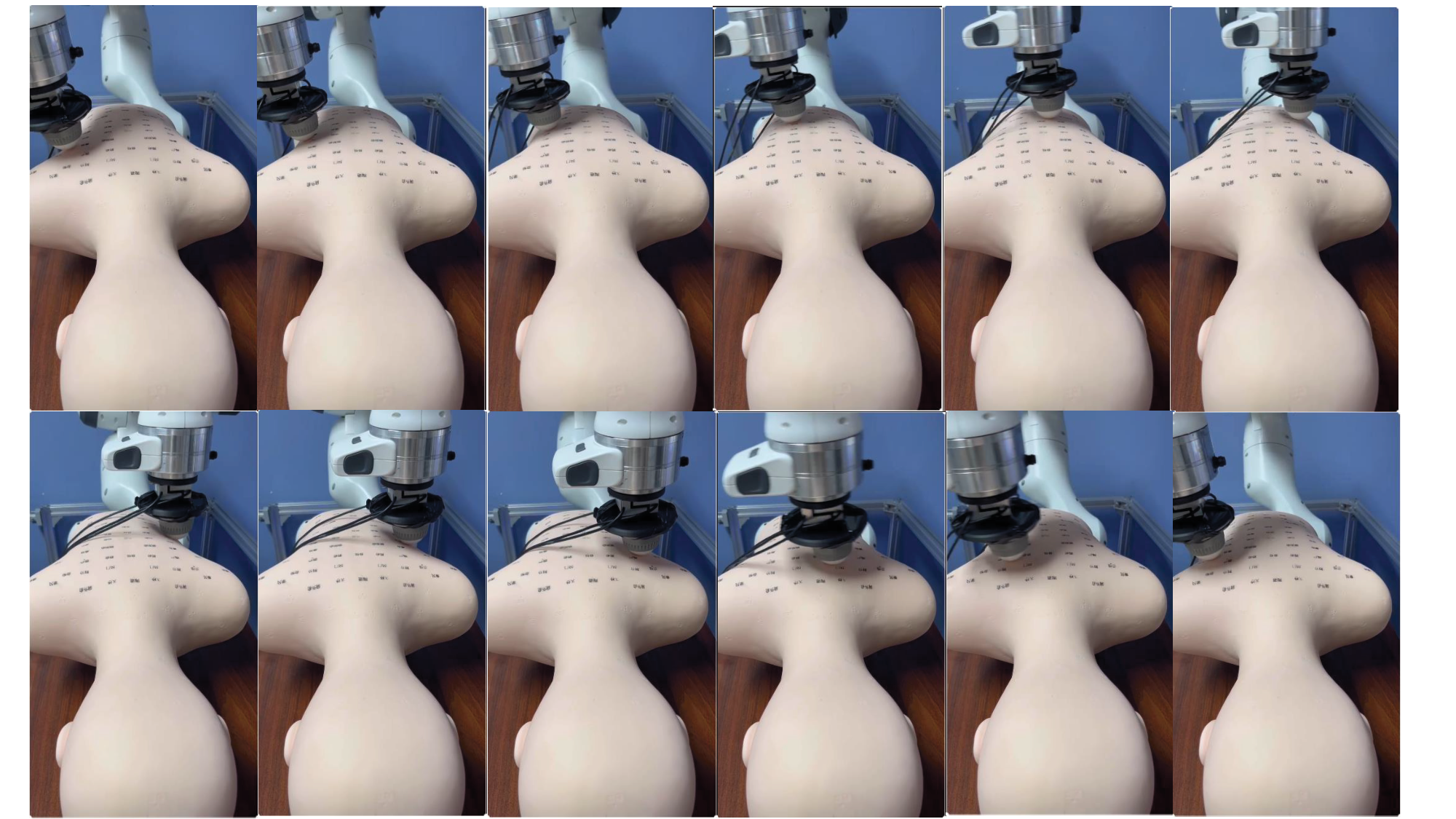} 
	\caption{
		Snapshots of the variable impedance control execution with the Franka Emika robot in the three-dimensional Cartesian space are presented. The robot performs a massage task on a silicone dummy, whose back is not planar. By utilizing the learned stiffness, the robot successfully completes the task.	}	
	\label{fig10}
\end{figure*}

In practical experiments, it is common for both the stiffness and damping parameters to be unknown and in need of estimation from demonstrations. Assuming the damping is an unknown constant, both the proposed algorithm and the convex optimization approach are used to estimate both the stiffness and damping. To handle the potential variation of damping over time. First, the stiffness and damping are estimated using the sliding windows. Then, the average damping value is calculated based on the estimated value obtained from each window. Subsequently, the stiffness is re-estimated using the calculated average damping. This two-step process allows for more accurate estimation of both the stiffness and damping and satisfies the constant damping assumption. The estimated results are shown in Fig. \ref{fig7}, which demonstrate that even without prior knowledge of the damping, the proposed algorithm and the Convex optimization method can accurately estimate both the stiffness and the damping. Although the convex optimization has a better performance,  notably, the proposed algorithm demonstrates a distinct advantage in terms of speed, as the two algorithms are implemented twice, making it considerably faster than the Convex optimization approach.

\subsection{Simulation for Validating the Novel Lyapunov Function}
To evaluate the effectiveness of the proposed novel   Lyapunov function, a series of simulations were conducted and  compared to the Lyapunov function proposed in \cite{stabilitytro} and a simple Lyapunov function. 

The section presents a reproduction of the simulation conducted in \cite{tank}, which involves a 1 degree of freedom (dof) system, 
\begin{equation}\label{equ:42}
	m\ddot{x}_{t}+ d_{t}\dot{x}_{t}+k_{t}{x}_{t}= 0,	
\end{equation}
with variable stiffness.

The reference trajectory and desired stiffness are defined as:
\begin{equation}\label{equ:43}
	\begin{cases}
		x^{d}_{t} = 10 \sin(0.1t),	\\ 
		k^{d}_{t} = k_c+10\sin(t),	
	\end{cases}	
\end{equation}
To validate the effectiveness of the proposed function, the damping is not designed as a constant but $2\varsigma \sqrt{k(t)} $, with $\varsigma$ being a positive constant to satisfy critical damping.
The system is first simulated using a mass $m$ of 10 Kg,  $k_c=15.5$ and damping $d_{t}=2\sqrt{k_t^d}$, then it is not always satisfy the stability conditions $0.5\dot{k}_t+0.5\alpha \dot{d}_t-\alpha k_t<0$ proposed by \cite{stabilitytro} but satisfy the stability conditions $0.5\dot{k}_t-\alpha k_t-0.25\alpha^3+0.25\alpha^2d_t<0$ proposed in this paper, where $\alpha=\min(d(t)/m)$ and the results are shown in Fig. \ref{fig8} and Fig. \ref{fig9}. 

\begin{figure}[htbp]
	\includegraphics[scale=0.8]{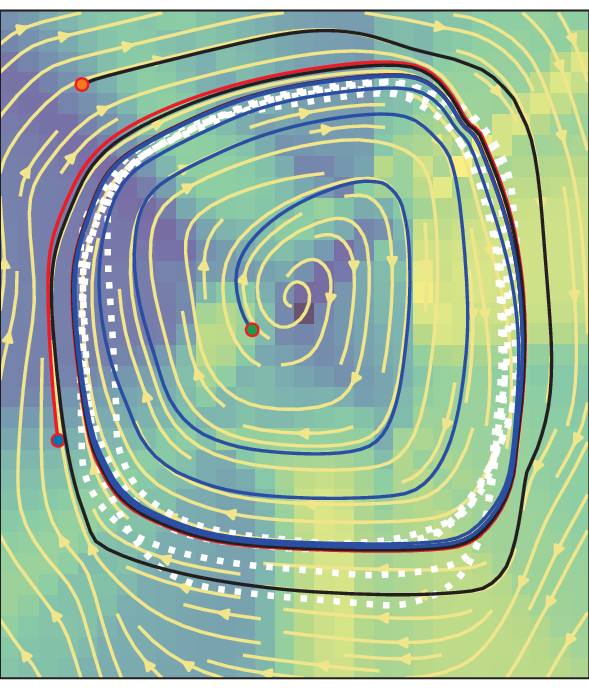} 
	\caption{The limit cycle is learned from the demonstration, with the white dotted line representing the demonstration trajectory. The black, red, and blue lines indicate trajectories that start from different random points and converge to the limit cycle.}
	\label{fig11}
\end{figure}
\begin{figure}[htbp]
	\includegraphics[scale=0.2]{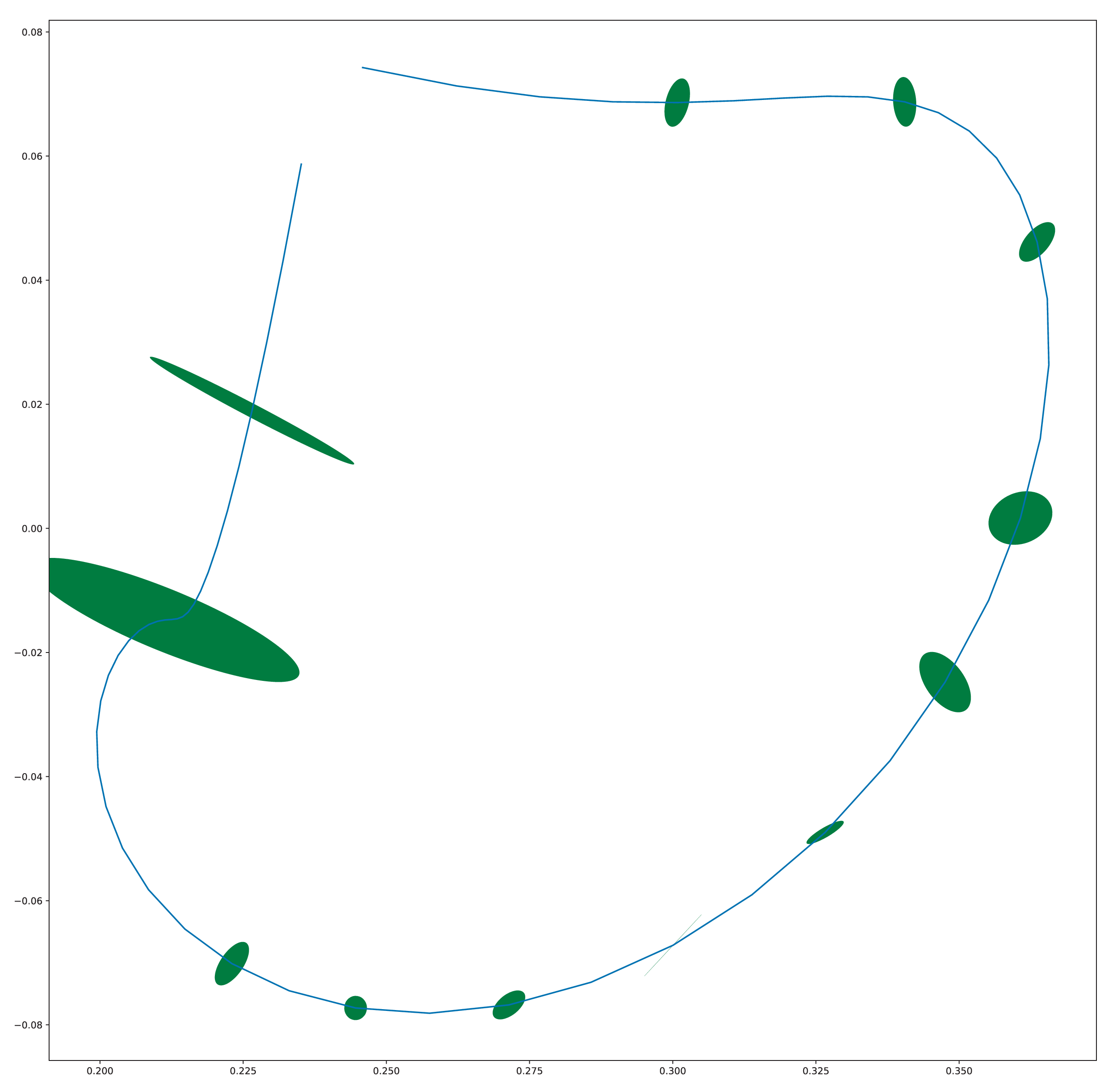} 
	\caption{The estimated stiffness is based on a segment of the demonstration trajectory, visualized using a stiffness ellipsoid.}
	\label{fig12}
\end{figure}
\begin{figure}[htbp]
	\includegraphics[scale=0.4]{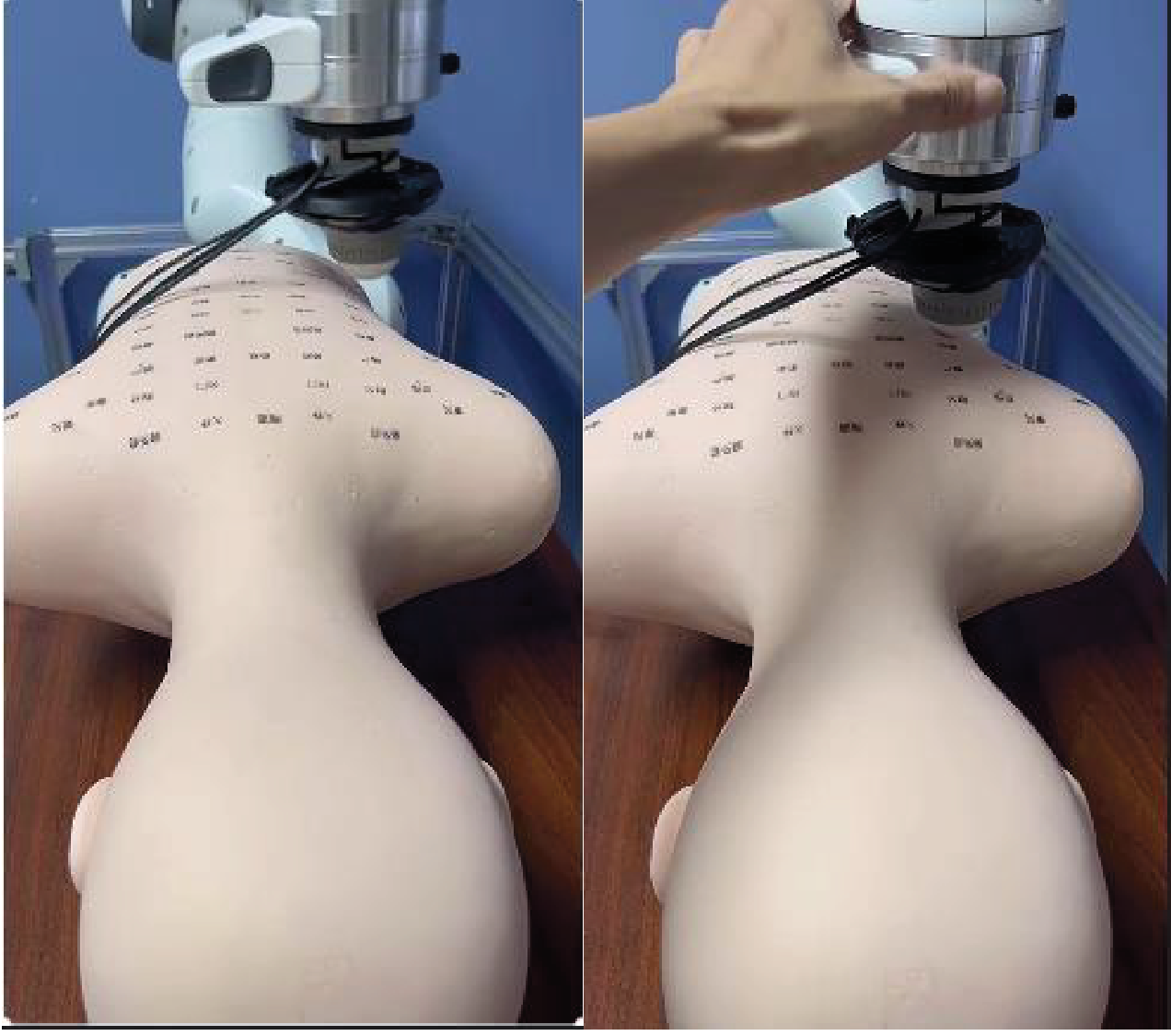} 
	\caption{Snapshots illustrate the deficiencies of using high and low stiffness for the massage task. With high stiffness (shown on the left), the massage ball may fail to conform to the surface of the silicone dummy, leading to ineffective massage. Conversely, with low stiffness (shown on the right), the robot may become trapped and require additional force to disengage, making it unsuitable for the massage task.}
	\label{fig13}
\end{figure}
It can be seen that with manually designed parameters, the Lyapunov function presented in this study, offers more flexible conditions than the counterpart proposed in \cite{stabilitytro}. Unlike the function in \cite{stabilitytro}, which can encounter local energy peaks that may breach necessary stability conditions, the proposed function ensures a monotonic and consistent reduction in energy over time. This characteristic guarantees not only robust stability but also a more enduring and reliable performance.

\subsection{ Validating  the Robustness of the Proposed Algorithm in Robot Massage Experiments}

To enable the robot to perform the massage task on a silicone dummy, a human initially guides the end effector of the Franka Emika robot to carry out the massage. During this demonstration, data is collected, including force measurements from the force sensor and motion information from the Franka robot. The trajectory information is used to learn a limit cycle using the algorithm proposed in \cite{mylimit}, thereby constructing a motion model.

The stiffness is then estimated using the motion model in conjunction with the demonstrated force data. It is evident that the demonstrated trajectory employs varying stiffness at different points because the back of the silicone dummy is not a flat surface. Humans naturally have the ability to adjust stiffness to perform massage tasks effectively while conforming to the skin's surface. 

Since the task is reproduced on the same silicone dummy and performed on the same back, the stiffness is modeled with respect to the position. Variable impedance control is applied, while a PID force control is used for vertical orientation. During reproduction, both the stiffness and the reference position and velocity are generated based on the current position. The damping is kept constant initially, and the stiffness is used to assess the system's passivity. If the system is not passive, the damping is adjusted to meet passivity constraints. 

To verify the effectiveness of the learned stiffness, the same massage task is performed using both constant high-stiffness and low-stiffness variable control. 
When using constant stiffness, the system always remains passive. However, with high stiffness, the robot sometimes loses its ability to conform to the surface of the silicone dummy's skin, particularly when the massage ball moves from a higher to a lower position. In contrast, with low stiffness, the robot maintains the ability to conform to the silicone dummy's skin surface. However, it may become trapped in the lower position and lack the necessary control torque to return, as the low stiffness does not provide sufficient control. With the learned variable stiffness, the robot can consistently conform to the silicone dummy's skin surface while effectively completing the massage task.
\section{Conclusions and Future Work}\label{sec5}
This paper proposes a framework for learning variable impedance control from demonstrations with guaranteed passivity. The performance of the proposed framework is evaluated using examples from manually designed datasets and a real robot massage task. The experimental results confirm the effectiveness of the method. However, the current tasks are limited to a fixed dummy, and future work will focus on incorporating variable end poses with real human subjects.



\begin{thebibliography}{1}

\bibitem{b1}	
Y. Zimmermann, J. Song, C. Deguelle, J. Laderach, L. Zhou, M. Hutter, R. Riener, and P. Wolf, "Human–Robot Attachment System for Exoskeletons: Design and Performance Analysis," \emph{ IEEE Transactions on Robotics,} vol. 39, no. 4, pp. 3087-3105, 2023.

\bibitem{b2}
A.C. Dometios and C.S. Tzafestas, "Interaction Control of a Robotic Manipulator With the Surface of Deformable Object," \emph{IEEE Transactions on Robotics,} vol. 39, no. 2, pp. 1321-1340, 2023.


\bibitem{b3}
H. Ravichandar, A.S. Polydoros, and S. Chernova,  "Recent Advances in Robot Learning from Demonstration," \emph{Annual Review of Control Robotics and Autonomous Systems,} vol. 3, no. 1, pp. 297-330, 2020.


\bibitem{b4}
L. Biagiotti, R. Meattini, D. Chiaravalli, G. Palli and C. Melchiorri, "Robot Programming by Demonstration: Trajectory Learning Enhanced by sEMG-Based User Hand Stiffness Estimation," \emph{ IEEE Transactions on Robotics,}  vol. 39, no. 4, pp. 3259-3278,  2023.

\bibitem{ralzy}
Y. Zhang, L. Cheng, H. Li and R. Cao, "Learning Accurate and Stable Point-to-Point Motions: A Dynamic System Approach," \emph{IEEE Robotics and Automation Letters}, vol. 7, no. 2, pp. 1510-1517, 2022.

\bibitem{b6}
C. Zeng, Y. Li, J. Guo, Z. Huang, N. Wang and C. Yang, "A Unified Parametric Representation for Robotic Compliant Skills With Adaptation of Impedance and Force," \emph{IEEE/ASME Transactions on Mechatronics,} vol. 27, no. 2, pp. 623-633, 2022.


\bibitem{b7}
T.K. Best, C.G. Welker, E.J. Rouse and R.D. Gregg, "Data-Driven Variable Impedance Control of a Powered Knee–Ankle Prosthesis for Adaptive Speed and Incline Walking,"  \emph{IEEE Transactions on Robotics,} vol. 39, no. 3, pp. 2151-2169, 2023.


\bibitem{b8}
E. Burdet, R. Osu, D. Franklin, T.E. Milner, and M. Kawato, "The central nervous system stabilizes unstable dynamics by learning optimal impedance," \emph{Nature,} vol. 414, pp. 446–449, 2001.

\bibitem{b9}
C. Yang, G. Ganesh, S. Haddadin, S. Parusel, A. Albu-Schaeffer and E. Burdet, "Human-Like Adaptation of Force and Impedance in Stable and Unstable Interactions," \emph{IEEE Transactions on Robotics,} vol. 27, no. 5, pp. 918-930,  2011.

\bibitem{b10}
Y. Michel, R. Rahal, C. Pacchierotti, P. R. Giordano and D. Lee, "Bilateral Teleoperation With Adaptive Impedance Control for Contact Tasks," \emph{IEEE Robotics and Automation Letters}, vol. 6, no. 3, pp. 5429-5436, 2021.

\bibitem{b11}
Y. Lin, Z. Chen and B. Yao, "Unified Motion/Force/Impedance Control for Manipulators in Unknown Contact Environments Based on Robust Model-Reaching Approach," \emph{ IEEE/ASME Transactions on Mechatronics,} vol. 26, no. 4, pp. 1905-1913,  2021.

\bibitem{b12}
R. Martín-Martín, M. A. Lee, R. Gardner, S. Savarese, J. Bohg and A. Garg, "Variable Impedance Control in End-Effector Space: An Action Space for Reinforcement Learning in Contact-Rich Tasks,"  \emph{in Proceedings ofIEEE/RSJ International Conference on Intelligent Robots and Systems}, Macau, China, 2019, pp. 1010-1017.

\bibitem{b13}
B. Jonas, S. Freek, T. Evangelos, and S. Stefan,  "Learning
variable impedance control," \emph{ The International Journal of
	Robotics Research,} vol. 30, no. 7, pp. 820-833, 2011.


\bibitem{b14}
Y. Zhu, Q. Wu, B. Chen and Z. Zhao, "Design and Voluntary Control of Variable Stiffness Exoskeleton Based on sEMG Driven Model," \emph{ IEEE Robotics and Automation Letters}, vol. 7, no. 2, pp. 5787-5794,  2022.

\bibitem{b15}
J. Arnold and H. Lee, "Variable Impedance Control for pHRI: Impact on Stability, Agility, and Human Effort in Controlling a Wearable Ankle Robot," \emph{ IEEE Robotics and Automation Letters}, vol. 6, no. 2, pp. 2429-2436, 2021.



\bibitem{stifftro}
L. Rozo, S. Calinon, D. G. Caldwell, P. Jimenez, and C. Torras, "Learning physical collaborative robot behaviors from human demonstrations,"
\emph{IEEE Transactions on Robotics,} vol. 32, no. 3, pp. 513-527, 2016.


\bibitem{ras}
F.J. Abu-Dakka, L. Rozo, and D. Caldwell, "Force-based variable
impedance learning for robotic manipulation," \emph{Robotics and Autonomous Systems,} vol. 109, pp. 156-167, 2018.


\bibitem{emg2}
C. Zeng, C. Yang, H. Cheng, Y. Li, and S.L. Dai, "Simultaneously
encoding movement and sEMG-based stiffness for robotic skill learning,"
\emph{IEEE Transactions on Industrial Informatics}, vol. 17, no. 2, pp. 1244–1252, 2021.

\bibitem{emgtm}
Y. Zhao, K. Qian, S. Bo, Z. Zhang, Z. Li, G. Li, A.A. Dehghani-Sanij, and S.Q. Xie, "Adaptive Cooperative Control Strategy for a Wrist Exoskeleton Using Model-Based Joint Impedance Estimation," \emph{IEEE/ASME Transactions on Mechatronics,} vol. 28, no. 2, pp. 748-757, 2023. 

\bibitem{myras}
Y. Zhang, L. Cheng, R. Cao, H. Li, and C. Yang, "A neural network based framework for variable impedance skills learning from demonstrations," \emph{Robotics and Autonomous Systems}, vol. 160, 2023.

\bibitem{rl}
M. Bogdanovic, M. Khadiv and L. Righetti, "Learning Variable Impedance Control for Contact Sensitive Tasks," \emph{ IEEE Robotics and Automation Letters}, vol. 5, no. 4, pp. 6129-6136, Oct. 2020.



\bibitem{stabilitytro}
K. Kronander and A. Billard, "Stability Considerations for Variable Impedance Control," \emph{ IEEE Transactions on Robotics}, vol. 32, no. 5, pp. 1298-1305,  2016.
\bibitem{tank}
F. Ferraguti, C. Secchi, and C. Fantuzzi, “A tank-based approach to
impedance control with variable stiffness,” \emph{in Proceedings of IEEE International Conference on Robotics and Automation,} Karlsruhe, Germany,
2013, pp. 4948-4953.

\bibitem{b24}
F.E. Tosun and V. Patoglu, "Necessary and sufficient conditions for the
passivity of impedance rendering with velocity-sourced series elastic
actuation," \emph{IEEE Transactions on Robotics}, vol. 36, no. 3, pp. 757-772,
2020.

\bibitem{regression}
C. Bishop, "Pattern Recognition and Machine Learning," Information Science and Statistics, New York: Springer, Vo. 1, Ch. 6, pp. 291-294, 2006.


\bibitem{ok87}
O. Khatib, "A unified approach for motion and force control of robot manipulators: The operational space formulation," \emph{IEEE Journal on Robotics and Automation,} vol. 3, no. 1, pp. 43-53, 1987.
\bibitem{spd}
J.H. Nicholas    "Computing a nearest symmetric positive semidefinite matrix," \emph{Linear Algebra and its Applications, 103,} vol. 103, pp. 103–118, 1988.

\bibitem{bhogan}
F.A. Mussa-Ivaldi, N. Hogan, and E. Bizzi, "Neural, mechanical, and geometric factors subserving arm posture in humans," \emph{The Journal of Neuroscience}, vol. 5, no. 10, pp. 2732-2743, 1987.

\bibitem{metric}
S. Jayasumana, R. Hartley, M. Salzmann, H. Li, and M. Harandi, "Kernel methods on riemannian manifolds with gaussian rbf kernels," \emph{IEEE Transactions on Pattern Analysis and Machine Intelligence}, vol. 37, no. 12, pp. 2464-2477, 2015.

\bibitem{29}
S. Jayasumana, R. Hartley, M. Salzmann, H. Li, and M. Harandi, "Kernel methods on riemannian manifolds with gaussian rbf kernels," \emph{IEEE Transactions on Pattern Analysis and Machine Intelligence}, vol. 37, no. 12, pp. 2464-2477, 2015.

\bibitem{mylimit}
Y. Zhang, H. Zhang, Y. Zou, H. Li, and L. Cheng, "Stabilizing Dynamic Systems through Neural Network Learning: A
Robust Approach," arXiv:2309.08849, 2024.

\end{thebibliography}
%

\newpage

\end{document}